\documentclass[conference]{IEEEtran}

\usepackage{ifthen}
\usepackage{fainekos-macros}

\usepackage{algorithm}
\usepackage[noend]{algpseudocode}
\usepackage{etoolbox}

\algrenewcommand\algorithmicindent{1.0em}

\usepackage{todonotes}

\makeatletter
\newcommand*{\algrule}[1][\algorithmicindent]{%
  \makebox[#1][l]{%
    \hspace*{.2em}% <------------- This is where the rule starts from
    \vrule height .75\baselineskip depth .25\baselineskip
  }
}

\newcount\ALG@printindent@tempcnta
\def\ALG@printindent{%
    \ifnum \theALG@nested>0% is there anything to print
    \ifx\ALG@text\ALG@x@notext% is this an end group without any text?
    \else
    \unskip
    \ALG@printindent@tempcnta=1
    \loop
    \algrule[\csname ALG@ind@\the\ALG@printindent@tempcnta\endcsname]%
    \advance \ALG@printindent@tempcnta 1
    \ifnum \ALG@printindent@tempcnta<\numexpr\theALG@nested+1\relax
    \repeat
    \fi
    \fi
}
\patchcmd{\ALG@doentity}{\noindent\hskip\ALG@tlm}{\ALG@printindent}{}{\errmessage{failed to patch}}
\patchcmd{\ALG@doentity}{\item[]\nointerlineskip}{}{}{} % no spurious vertical space
\makeatother

\usepackage{graphicx}
\graphicspath{{figures/}}
\usepackage{subcaption}

\usepackage{caption}
\usepackage{float}

\usepackage{color,soul}
\newboolean{HIGHCOMM}
\setboolean{HIGHCOMM}{false}
\newcommand \yhl[1]{\ifthenelse{\boolean{HIGHCOMM}}{\textcolor{blue}{#1}}{#1}}
\newboolean{SHOW_WHAT_IT_WAS}
\setboolean{SHOW_WHAT_IT_WAS}{true}
\newcommand \shl[1]{\ifthenelse{\boolean{SHOW_WHAT_IT_WAS}}{\yhl{\sout{#1}}}{}}
\usepackage{ulem}
\usepackage{tikz}
\newboolean{TECHREP}
\setboolean{TECHREP}{true}

\begin{document}

\title{Extended LTLvis Motion Planning Interface\\
\Large (Extended Technical Report)}

\author{\IEEEauthorblockN{Wei Wei, Kangjin Kim and Georgios Fainekos}
\IEEEauthorblockA{The School of Computing, Informatics and Decision Systems Engineering
\\
Arizona State University\\
Tempe, AZ, USA\\
Email: \{wwei17,Kangjin.Kim,fainekos\}@asu.edu}}

\maketitle

\begin{abstract}
%This paper introduces an extended version of the Linear Temporal Logic (LTL) graphical interface.
This paper introduces a graphical interface for Linear Temporal Logic (LTL) specifications for mobile robots.
It is a sketch based interface built on the Android platform which makes the LTL control interface more friendly to non-expert users.
By predefining a set of areas of interest, this interface can quickly and efficiently create plans that satisfy extended plan goals in LTL.
The interface can also allow users to customize the paths for this plan by sketching a set of reference trajectories.
Given the custom paths by the user, the LTL specification and the environment, the interface generates a plan balancing the customized paths and the LTL specifications.
We also show experimental results with the implemented interface.
\end{abstract}


\IEEEpeerreviewmaketitle

\section{Introduction}

%There is a constant progress in robotics every day.
%Humans are stepping into to a new society - a society in which robots play significant roles in human daily life.
%Tasla Factory uses robots to assemble car parts%
%\footnote{http://bgr.com/2016/02/23/tesla-releases-awesome-time-lapse-videos-of-robots-assembling-a-model-x/};
%Amazon deploys robots to manage their warehouse%
%\footnote{http://www.chonday.com/Videos/how-the-amazon-warehouse-works};
%a lot of people buy home cleaning robots to clean their carpets.
%From industrial machines to housing robots, the ability of robots to accomplish complex tasks is increasing at a high rate.
%In the development of these robots, experts need to address not only the hardware design, but the control software as well.
%Robots designed for different purpose may require different control methods.
%Let us consider few control methods: To order a robot to move forward, a forward button is sufficient.
%If the robot is moving in a 2D plane, a joystick can be a good option.
%If the robot is required to reach a certain destination autonomously, a touch screen is a much easier option to locate the coordinates on a displayed map.
%Lastly, if there is also path that the robot should follow, a sketch interface should be handy to define this requirement.

As the robots become more capable, so does the need to specify and monitor complex motion and mission plans.
Temporal logics have been proposed as an effective specification language for complex missions for single \cite{FainekosGKP2009} and multiple \cite{SmithTBR2010} robots.
However, temporal logic specifications are not easy to write for people without extensive training in formal logic.
Therefore, in \cite{SrinivasKKKF2013}, we developed a graphical interface for Linear Temporal Logic (LTL) specifications.
In LTLvis, the user creates a graph structure in the workspace of the robot which is then translated into an LTL formula which is forwarded to the planner.

In this paper, we extend our work by allowing the user to incorporate specific path recommendations for certain parts of the mission.
In particular, we enable the user to sketch path segments on the user interface which are then taken into account by the planner.
Since we focus on supervised autonomy, the sketched paths are not trajectories to be tracked by the robot but rather additional constraints for the LTL planner.
Two challenges arise when adding such path constraints.
First, how to identify which path on the roadmap is the closest to the one sketched by the user.
Second, how to guarantee that the sketched path does not violate any other requirements provided as part of the LTL requirement.
In this paper, we provide algorithmic answers to both problems.
Furthermore, we demonstrate our framework using an iRobot Create (TurtleBot) and the LTL planning framework by \cite{UlusoyMB2013}.

%In order to fulfill these control requirements, 

\textbf{Related Work:} 
\yhl{
Graphical control interfaces appear to be an effective way to control mobile robots \cite{ShahSC2010}.
%are the most straightforward and easy to use for both expert and non-expert users \cite{ShahSC2010}. 
%In order to fulfill the varieties of these control requirements, graphic control interfaces are the most straightforward and easy to use for both expert and non-expert users \cite{ShahSC2010}. 
%\cite{SakamotoHII2009}.
With a graphical interface, users can control multiple robots more conveniently \cite{OchiaiTITO2014} by clicking a predefined button instead of writing a robot control program. 
The proposed work from \cite{OchiaiTITO2014} is similar to our approach.
%However, instead of commanding a robot to follow a path, they assign a start and a position for the robot and the robot will explore its path by searching the RRT of the given map.
However, instead of commanding a robot to follow a path, they assign a start position for the robot and the robot will explore the given map to find its path by searching Rapidly exploring Random Trees (RRT).
}

In \cite{SkubicBBAM2004}, the authors propose a methodology to extract spatial information about the sketched map and path.
This information including qualitative path movement, the key turning point of the path and high level path description is helpful to model the human-like robot navigation.
In \cite{FrankK2015}, the authors had shown that planning using sketch based interfaces can be improved using path correction.
%The interface is designed to be friendly to all users including the ones with Parkinson disease.
Once users draw a path bypassing an invalid region (collisions), this interface will auto-correct the invalid sub-path to a valid B\'ezier curve.
%It can also correct multiple collisions in the same path.
%Most of home cleaning robot users are non-expert programmers.
Sakamoto et al. proposed a robot control interface especially for home robots \cite{SakamotoHII2009}.
%In their work, a vacuum machine was controlled by the proposed interface to complete a set of tasks.
The authors define a set of gesture commands for a set of actions.
They include move with an open curve, vacuum with a closed curve, stop with a cross mark, etc.

%The work presented in \cite{LiuSII2011} proposed an interface solution to control multiple robots. In the interface design, users are able to give command to each individual robot in each individual
%interface layer.
%In our interface, when a user sketch crosses over an undefined region, the interface can still find its Best Matching Path (BMP).
%However, if this BMP conflicts with the LTL specification, it will be ignored without noticing the user. The interface also inherits all the gesture language from LTLvis \cite{SrinivasKKKF2013}. 

In terms of LTL planning, in \cite{SmithTBR2010}, the authors proposed a solution to generate the optimal plan under a temporal logic specification.
LTL is the high level specification for the planning task which is required to be repeatedly satisfied.
To let the robot complete the mission in a dynamic environment, Ulusoy et al. proposed a solution in \cite{UlusoyMB2013}.
As the robot sensors have limited ability to scan the whole environment, they define a limited region as the local environment. 

\textbf{Summary of Contributions:} The main contribution in our research is to combine an easy-to-use sketch-based interface with the expressive power of LTL and to improve the LTL path planner provided by \cite{UlusoyMB2013} for this hybrid interface%
\footnote{The authors in \cite{UlusoyMB2013} provide software package RHTL which includes
LTL planner (LOMAP). Our implementation is based on their software package.}.
%The whole system (interface and planner) aims to solve a more complex path planning problem for a single robot.
%Follow up work will concentrate to extending the curve results to multiple robots and on performing usability studies.
\yhl{
A secondary contribution, which is important on its own, is that we provide a greedy  algorithm to identify the closest path on a directed topologically grounded graph to a hand drawn curve. We remark that our algorithm allows the path to be cyclic.
}

\section{Preliminary}
In this section, we will first cover the graphical language for LTL. Then, we will review LTL path planning.

\subsection{Graphical Language for LTL}
Temporal logic is a logic that describes events in time.
%Linear Temporal Logic (LTL) is a modal temporal logic reasoning over an infinite sequence of states \cite{FainekosKP2005}.
Linear Temporal Logic (LTL) is a modal temporal logic reasoning over an infinite sequence of states.
This section mainly introduces the research work by Srinivas, et al on defining a graphical language \cite{SrinivasKKKF2013}.
\ifthenelse {\boolean{TECHREP}}
{
In particular, \cite{SrinivasKKKF2013} provides a graphical representation of an LTL formula in a 2D space.
The graph G is a tuple $(V, E, v_0, c, L, \Lambda, x)$:
\begin{itemize}
\item $V$ is the set of nodes;
\item $E \subseteq V \times V$ is the set of edges;
\item $v_0 \in V$ is the start node;
\item $c: V \rightarrow \{green, red\}$ is a function that colors each node either green or red, which corresponds to visiting or avoiding a node%
\footnote{Icons can be added to help people with color blindness.};
\item $L: V \rightarrow \Pi_{\BUCHI}(\tau)$ labels each node with an LTL formula over the set of propositions $\Pi$;
\item $\Lambda: E \rightarrow BO_1 \times BO_2 \times TO_2 \times TO_1$ is a function that labels each edge on the graph with one or more Boolean or temporal operators:
\begin{itemize}
\item $BO_1 = \{AND, OR\}$;
\item $BO_2 = BO_1 \cup \{\epsilon%
\footnote{$\epsilon$ denotes an empty symbol.}
, IMPLIES\}$;
\item $TO_1 = \{\epsilon, FUTURE, ALWAYS\}$;
\item $TO_2 = TO_1 \cup \{NEXT, UNTIL\}$
\end{itemize}
\item $x: V \rightarrow \mathbb{R}^{2}$ is the position of the node on the map or on the image
\end{itemize}
}
{
In particular, \cite{SrinivasKKKF2013} provides a graphical representation of an LTL formula in a 2D space. The graph G is a tuple $(V, E, v_0, c, L, \Lambda, x)$: where $V$ is the set of nodes; $E \subseteq V \times V$ is the set of edges; $v_0 \in V$ is the start node; $c: V \rightarrow \{green, red\}$ is a function that colors each node either green or red, which corresponds to visiting or avoiding a node%
\footnote{Icons can be added to help people with color blindness.};
$L: V \rightarrow \Pi_{\BUCHI}(\tau)$ labels each node with an LTL formula over the set of propositions $\Pi$;
$\Lambda: E \rightarrow BO_1 \times BO_2 \times TO_2 \times TO_1$ is a function that labels each edge on the graph with one or more Boolean or temporal operators 
(in detail,
$BO_1 = \{AND, OR\}$;
$BO_2 = BO_1 \cup \{\epsilon%
\footnote{$\epsilon$ denotes an empty symbol.}
, IMPLIES\}$;
$TO_1 = \{\epsilon, FUTURE, ALWAYS\}$;
$TO_2 = TO_1 \cup \{NEXT, UNTIL\}$);
$x: V \rightarrow \mathbb{R}^{2}$ is the position of the node on the map or on the image.
}

%\todo[inline]{TODO: Add Figure 1 for all possible combinations of $\Lambda$.}
\begin{figure}
\centering
\includegraphics[width=0.35\textwidth]{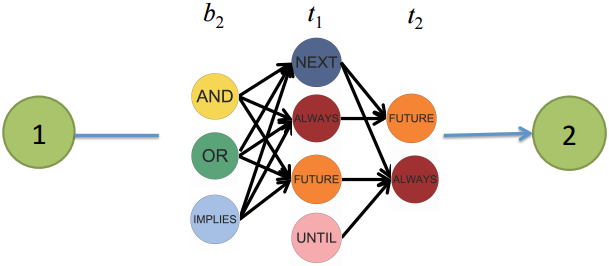}
\caption{The allowed combination of Boolean and temporal operators over an edge (Reproduced from \cite{SrinivasKKKF2013})}
\label{fig:combination}
\end{figure}

As $BO_1$ is always implicitly used to connect consecutive propositions, it is not included when forming the graph.
Figure \ref{fig:combination} is the flowchart of possible values of $\Lambda$.
%\todo[size=\tiny,inline]{We note that :\\
%1. \textcircled{1} and \textcircled{2} are vertices in $V$, and \\
%2. $\epsilon$ can make the label over the edge between \textcircled{1} and \textcircled{2} empty.}

\subsection{LTL Path Planning}
Path planning is the problem of finding a path between a start position and an end position.
Temporal logic path planning is the path planning problem whose result, i.e., path must satisfy a
temporal logic requirement.
%The basic theory on temporal logic planning is described in  \cite{FainekosKP2005,FainekosGKP2009,SmithTBR2010}.
The basic theory on temporal logic planning is described in  \cite{FainekosGKP2009,SmithTBR2010}.
First we need to represent an environment as a discrete graph.
%Here, we introduce the definition of transition system.
\ifthenelse {\boolean{TECHREP}}
{
\begin{defn}(TS) A transition system is a tuple
$T := (Q_{TS}, q_{init}, \delta_{TS}, \Pi, h, w_{TS})$, where $Q_{TS}$ is a set of
states.
It represents the accessible area in the graph;
\begin{itemize}
\item $q_{init} \in Q_{TS}$ is the starting state;
\item $\delta_{TS} \subseteq Q_{TS} \times Q_{TS}$ denotes the transition relation between two states;
\item $\Pi$ is a finite set of atomic propositions; 
\item $hℎ: Q_{TS} \rightarrow 2^{\Pi}$ is a function labeling areas in the environment with atomic propositions;
\item $w_{TS} : \delta \rightarrow \mathbb{N}$ is
the weight assigned to each transition.
\end{itemize}
\end{defn}
}
{
\begin{defn}(TS) A transition system is a tuple
$T := (Q_{TS}, q_{init}, \delta_{TS}, \Pi, h, w_{TS})$, where $Q_{TS}$ is a set of
states. It represents the accessible area in the graph; $q_{init} \in Q_{TS}$ is the starting state; $\delta_{TS} \subseteq Q_{TS} \times Q_{TS}$ denotes the
transition relation between two states; $\Pi$ is a finite set of
atomic propositions; $hℎ: Q_{TS} \rightarrow 2^{\Pi}$ is a function labeling areas
in the environment with atomic propositions; $w_{TS} : \delta \rightarrow \mathbb{N}$ is
the weight assigned to each transition.
\end{defn}
}

We denote a finite path on the transition system as $p = q_0,q_1,\dots,q_n$, where $q_0 = q_{init}$ and for $0 \leq k < n$, $q_k \in Q_{TS}$ and $(q_k , q_{k+1}) \in \delta$.
The result generated from running this path is a word $v_0 v_1 \dots$, where $v_k = h(q_k)$ is the set of atomic propositions
satisfied at $q_k$.

%After discretizing the environment into a transition system, we also need to convert the specification into the same format.
After transferring a given environment into a discretized transition system TS, we also need to convert a given LTL specification.
Thanks to the tool %
%\footnote{LTL2BA is a tool accepting LTL formulas as input and returning a B\"{u}chi
%automaton as output.}
provided by \cite{GastinO2001}, we can easily convert any LTL formula into a B\"{u}chi automaton. We introduce the definition of a B\"{u}chi automaton.

\ifthenelse {\boolean{TECHREP}}
{
\begin{defn}(BA) A B\"{u}chi automaton is a tuple $\BUCHI := (Q_{BA}, Q_{init}, \delta_{BA}, \Sigma, F_{BA})$, where $Q_{BA}$ is a set of states;
\begin{itemize}
\item $Q_{init}$ is a set of initial states; 
\item $\delta_{BA} \subseteq Q_{BA} \times \Sigma \times Q_{BA}$ is a transition relation;
\item $\Sigma$ is the input alphabet;
\item $F_{BA}$ is a set of
accepting states.
\end{itemize}
\end{defn}
}
{
\begin{defn}(BA) A B\"{u}chi automaton is a tuple $\BUCHI := (Q_{BA}, Q_{init}, \delta_{BA}, \Sigma, F_{BA})$, where $Q_{BA}$ is a set of states;
$Q_{init}$ is a set of initial states; $\delta_{BA} \subseteq Q_{BA} \times \Sigma \times Q_{BA}$ is a transition relation; $\Sigma$ is the input alphabet; $F_{BA}$ is a set of
accepting states.
\end{defn}
}

For a run of input word $W = \omega_0 \omega_1 \dots$ on the B\"{u}chi automaton
where $\omega_i \in \Sigma$, the resulting sequence would be $r =
s_0 s_1 \dots$, where $s_i \in Q_{BA}$ and $(s_i, \omega_i, s_{i+1}) \in \delta_{BA}$.
Now we have both TS and BA in a graph format.
The goal is to find a resulting sequence $r = c_0 c_1 \dots$ where $c_i := (q_j, s_k)$, $q_j \in Q_{TS}$ and $s_k \in Q_{BA}$.
The resulting sequence should be valid in TS and ending at one accepting state in BA%
\footnote{For an infinite word, the word should contain at least one accepting state in BA infinitely often.}.
Hence, we need to construct a product automaton $P := TS \times BA$.

\ifthenelse {\boolean{TECHREP}}
{
\begin{defn}(PA) The product automaton $P = TS \times BA$ between the transition system $T$$:=$$(Q_{TS},$$q_{init},$$\delta_{TS},$$\Pi,$$h,$$w_{TS})$ and B\"uchi automaton $\BUCHI$$:=$$(Q_{BA},$$Q_{init},$$\delta_{BA},$$\Sigma,$$F_{BA})$ is a tuple $P$$:=$$(S_P,$$S_{PO},$$\delta_P,$$w_P,$$F_P)$, where $S_P$$=$$Q_{TS}$$\times$$Q_{BA}$ is a finite set of states;
\begin{itemize}
\item $S_{PO} = q_{init} \times Q_{init}$ is the set of initial states;
\item $\delta_P \subseteq \delta_{TS} \times \delta_{BA}$ is a transition relation and $((q_i, s_i), (q_j, s_j)) \in \delta_P$ if and only if $(q_i, q_j) \in \delta_{TS}$ and $(s_i, \omega_i, s_j) \in \delta_{BA}$;
\item $w_P((q_i,s_i),(q_j,s_j))$$=$$w_{TS}(q_i,q_j)$ is a weight function;
\item $F_P$$=$$Q_{TS}$$\times$$F_{BA}$ is a set of accepting states.
\end{itemize}
\end{defn}
}
{
\begin{defn}(PA) The product automaton $P = TS \times BA$ between the transition system $T$$:=$$(Q_{TS},$$q_{init},$$\delta_{TS},$$\Pi,$$h,$$w_{TS})$ and B\"uchi automaton $\BUCHI$$:=$$(Q_{BA},$$Q_{init},$$\delta_{BA},$$\Sigma,$$F_{BA})$ is a tuple $P$$:=$$(S_P,$$S_{PO},$$\delta_P,$$w_P,$$F_P)$, where $S_P$$=$$Q_{TS}$$\times$$Q_{BA}$ is a finite set of states; $S_{PO} = q_{init} \times Q_{init}$ is the set of initial states; $\delta_P \subseteq \delta_{TS} \times \delta_{BA}$ is a transition relation and $((q_i, s_i), (q_j, s_j)) \in \delta_P$ if and only if $(q_i, q_j) \in \delta_{TS}$ and $(s_i, \omega_i, s_j) \in \delta_{BA}$; $w_P((q_i,s_i),(q_j,s_j))$$=$$w_{TS}(q_i,q_j)$ is a weight function; $F_P$$=$$Q_{TS}$$\times$$F_{BA}$ is a set of accepting states.
\end{defn}
}

The set of final states $F_P$ of the product automaton represents the ultimate goal of the planning path. Then we can reduce the problem of LTL path planning into finding the optimal path on a graph given a starting position.
At this level, many methods can be utilized such as A$^{*}$, DFS, Dijkstra etc. For example, if
the resulting path is $(q_0, s_0)$, $(q_1, s_1)$, $\dots$, $(q_n, s_n)$, then the actual path on the transition system (robot workspace) will be $q_0$, $q_1$, $\dots$, $q_n$.

\section{Problem Description}
%We will describe our problem and the expected solution.
\subsection{Problem Overview}
This research mainly focuses on the problems of solving the path planning under a given LTL specification.
Given an environment, a graphical LTL specification, and the user's preferred paths sketched on the environment, find the optimal path satisfying the LTL specification and maximally following the user's path sketches.
Once there exist conflicts between the user sketch path and the LTL specification, the interface should be able to regard the LTL specification as a higher priority requirement and find an alternative path minimizing the distance from the user sketch path.
The rationale behind this choice is that the user may not be explicitly aware of important safety requirements and event dependencies when drawing the desired path.
An alternative approach would be to recommended revisions to the mission requirements based on the path drawn by the user.
We have contacted similar research in the past in the context of LTL planning under user preferences \cite{KimG2014}.

\subsection{Solution Overview}
The interface starts with an empty screen asking the user to input a map image.
Then, the user can sketch on the map using the interface.
There are three different editing modes for planning, roadmap editing, and LTL editing.

%\todo[inline]{TODO:Add Figures 2, 5 and so on.}
\ifthenelse {\boolean{TECHREP}}
{
\begin{figure}
\centering
\begin{subfigure}[b]{0.24\textwidth}
\centering
\includegraphics[width=\textwidth]{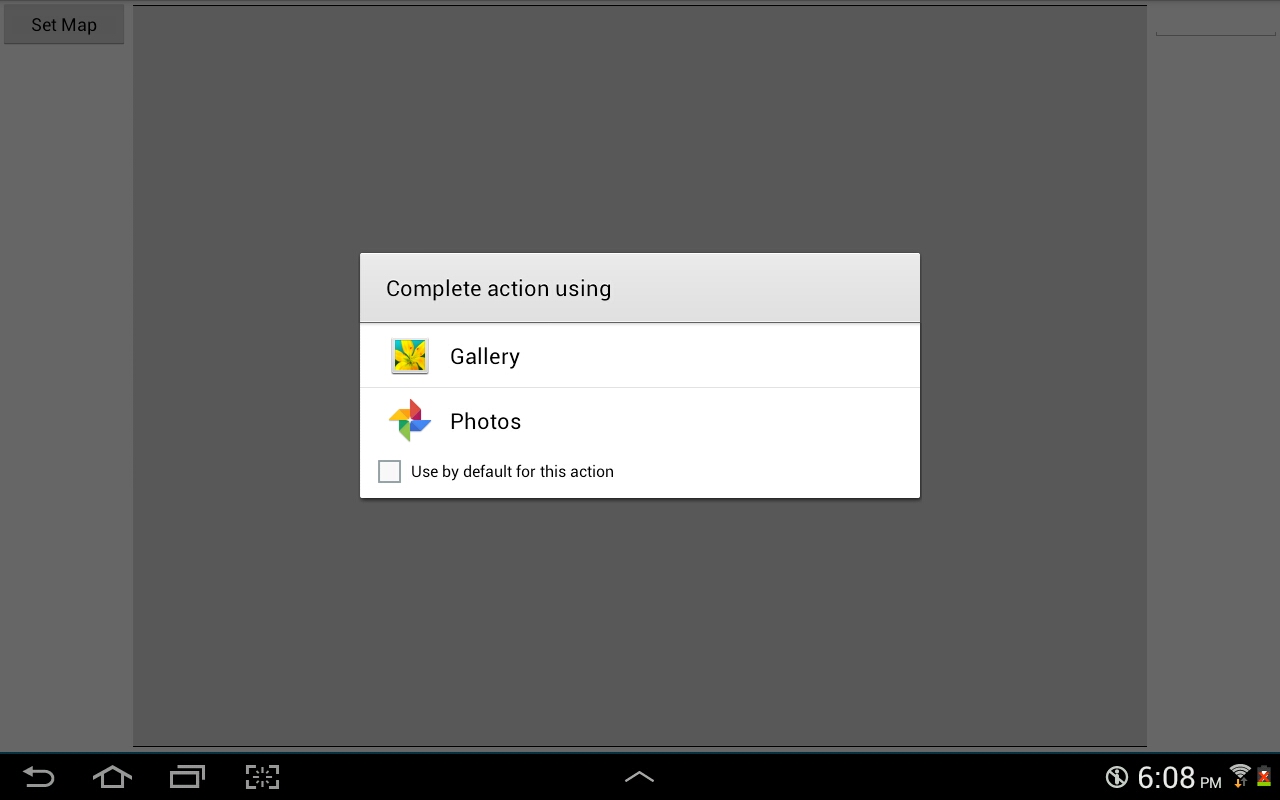}
\end{subfigure}
%\hfill
\begin{subfigure}[b]{0.24\textwidth}
\centering
\includegraphics[width=\textwidth]{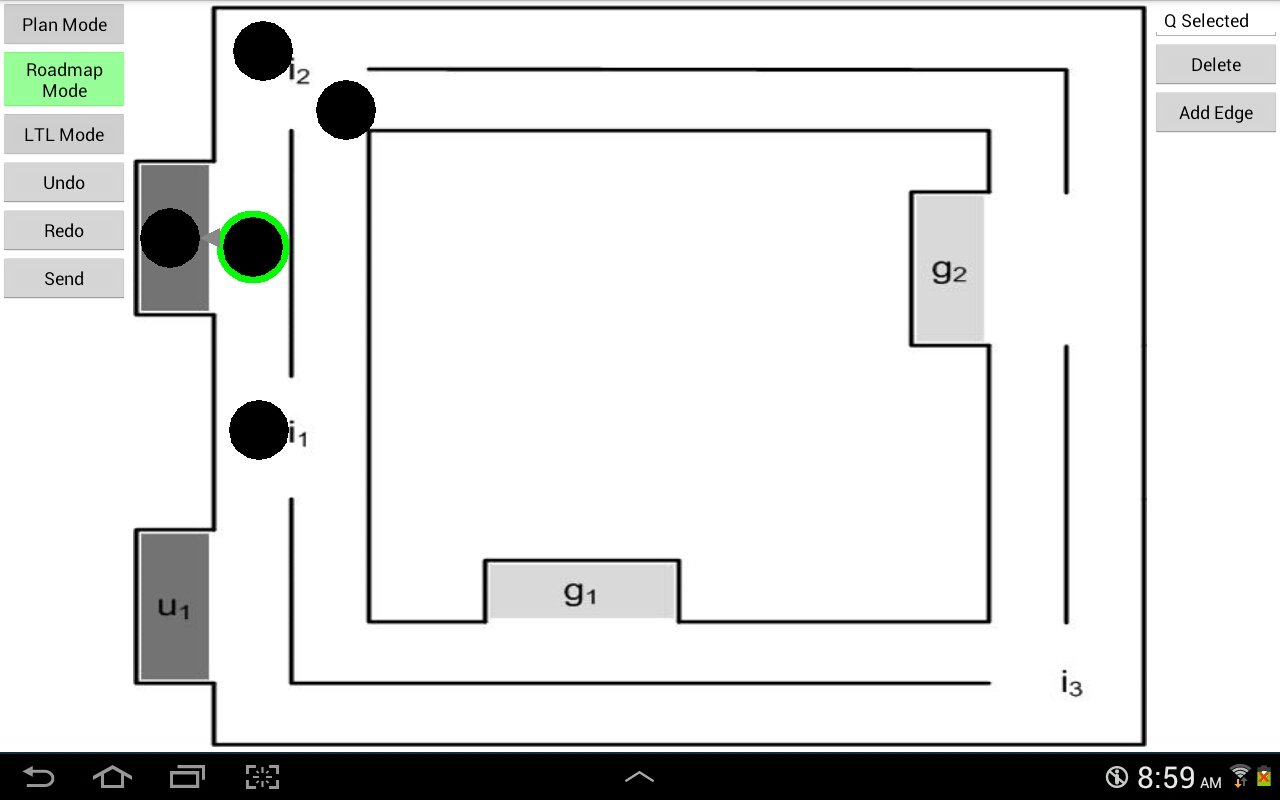}
\end{subfigure}
\vskip\baselineskip
\begin{subfigure}[b]{0.24\textwidth}
\centering
\includegraphics[width=\textwidth]{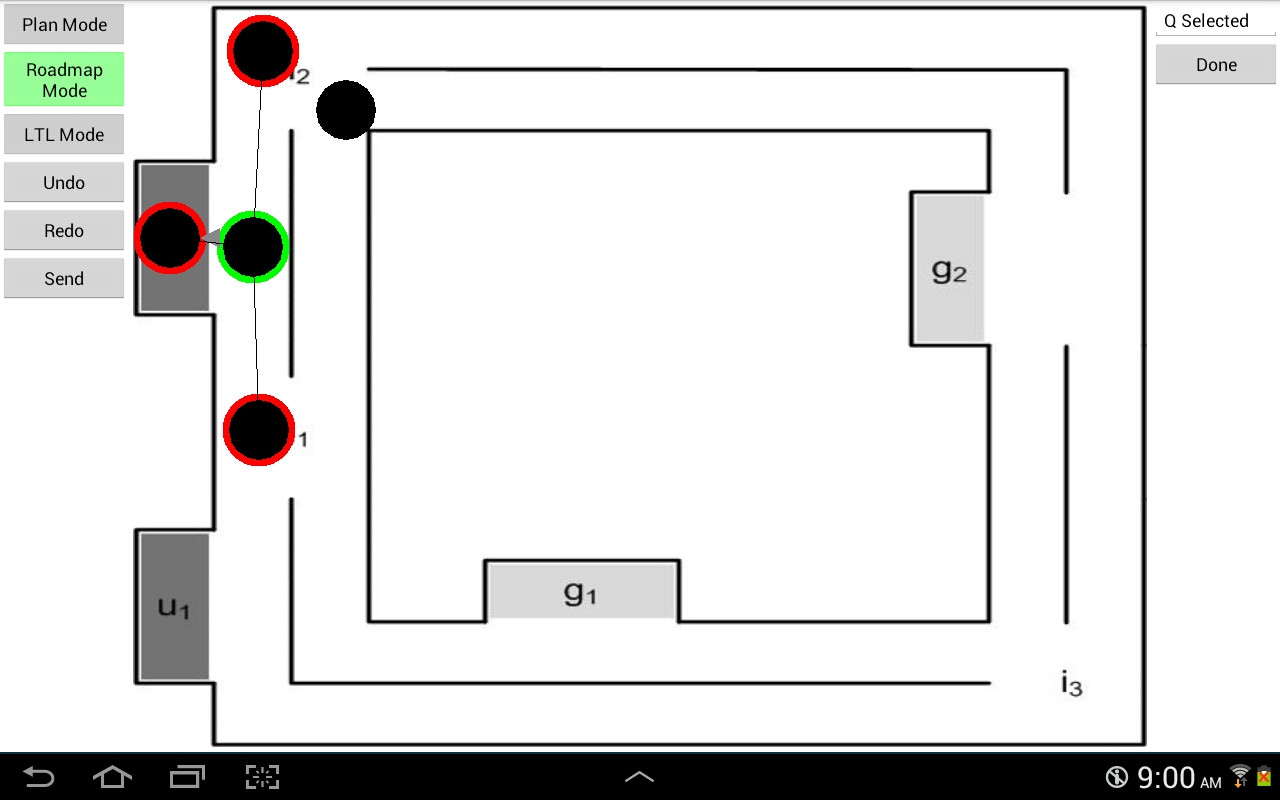}
\end{subfigure}
%\hfill
\begin{subfigure}[b]{0.24\textwidth}
\centering
\includegraphics[width=\textwidth]{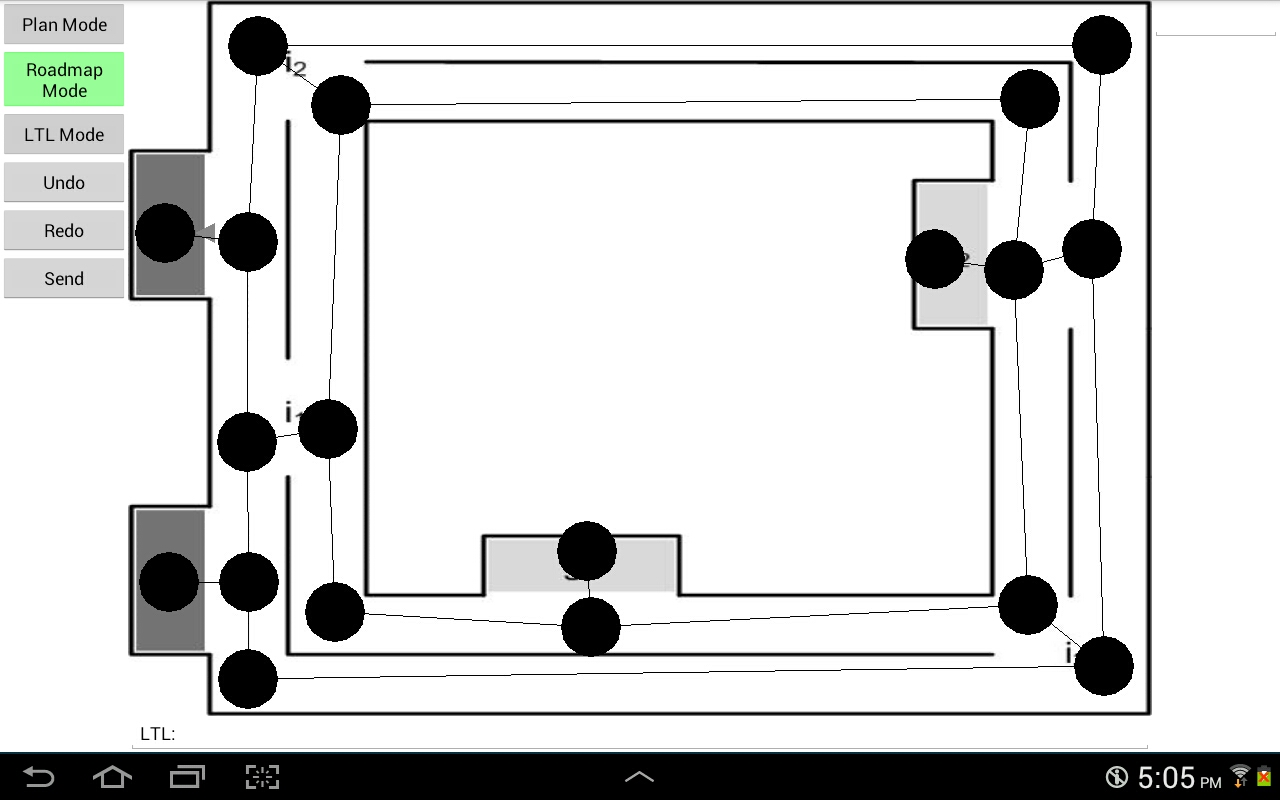}
\end{subfigure}
\caption{The procedure of manually creating a roadmap.
Upper left: the user is asked to load a map.
Upper right: when a node is selected, it will be colored green.
Lower left: the red nodes are denoted as neighbors of the green node.
Lower right: a complete roadmap.}
\label{fig:manual_procedure}
\end{figure}
}
{
\begin{figure}
\centering
\begin{subfigure}[b]{0.2\textwidth}
\centering
\includegraphics[width=\textwidth]{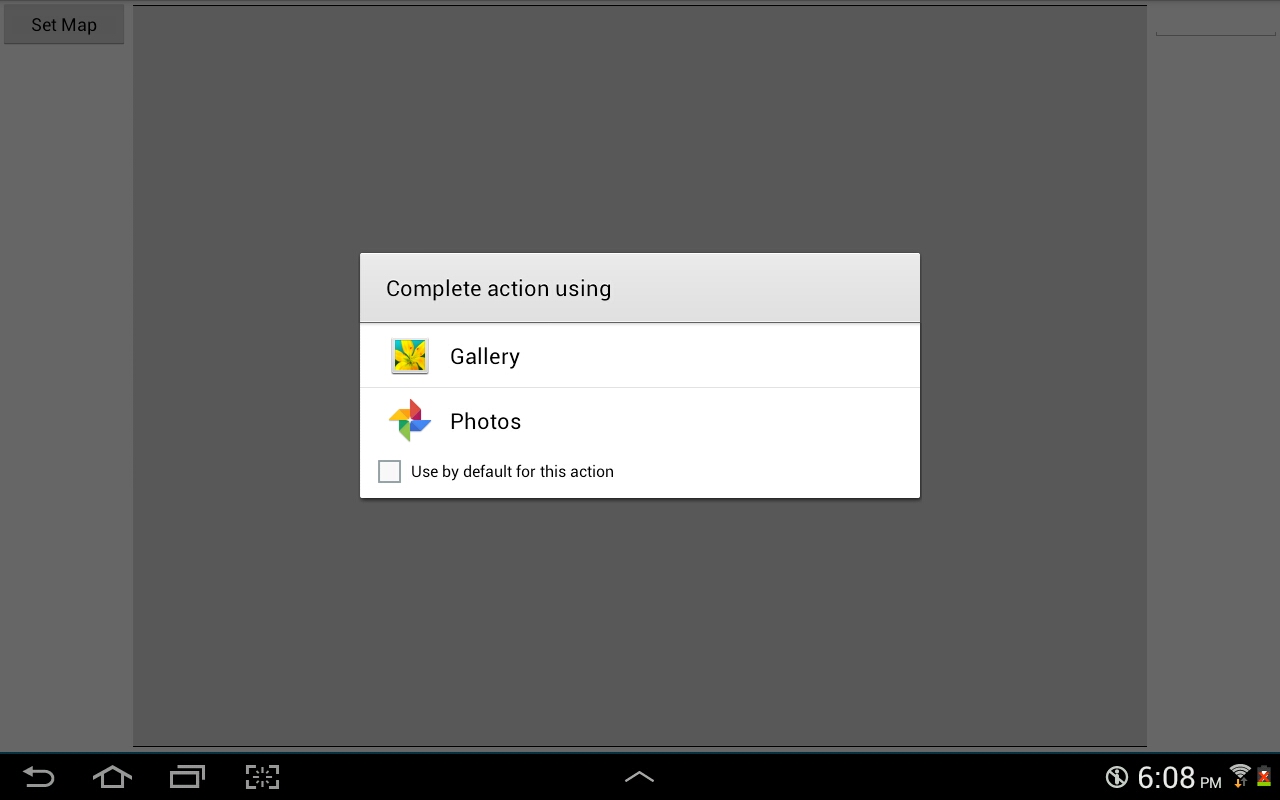}
\end{subfigure}
%\hfill
\begin{subfigure}[b]{0.2\textwidth}
\centering
\includegraphics[width=\textwidth]{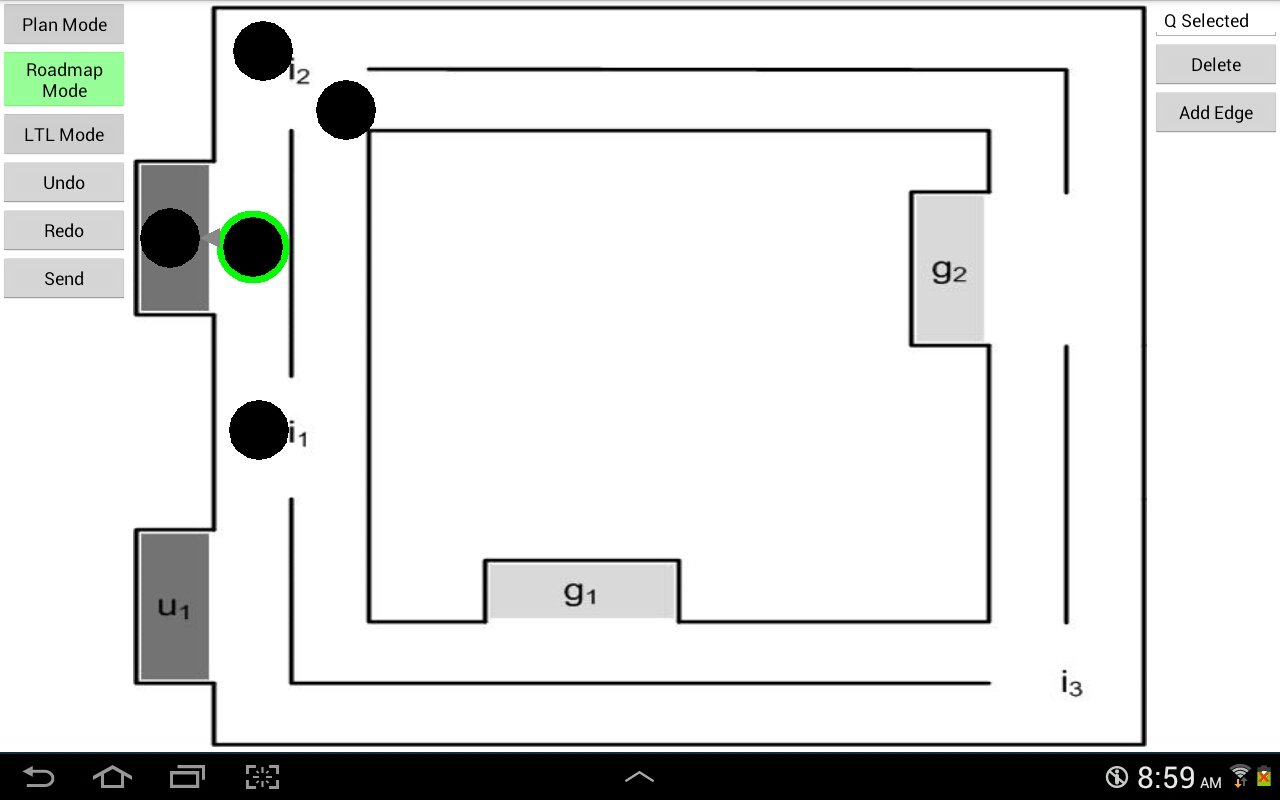}
\end{subfigure}
\vskip\baselineskip
\begin{subfigure}[b]{0.2\textwidth}
\centering
\includegraphics[width=\textwidth]{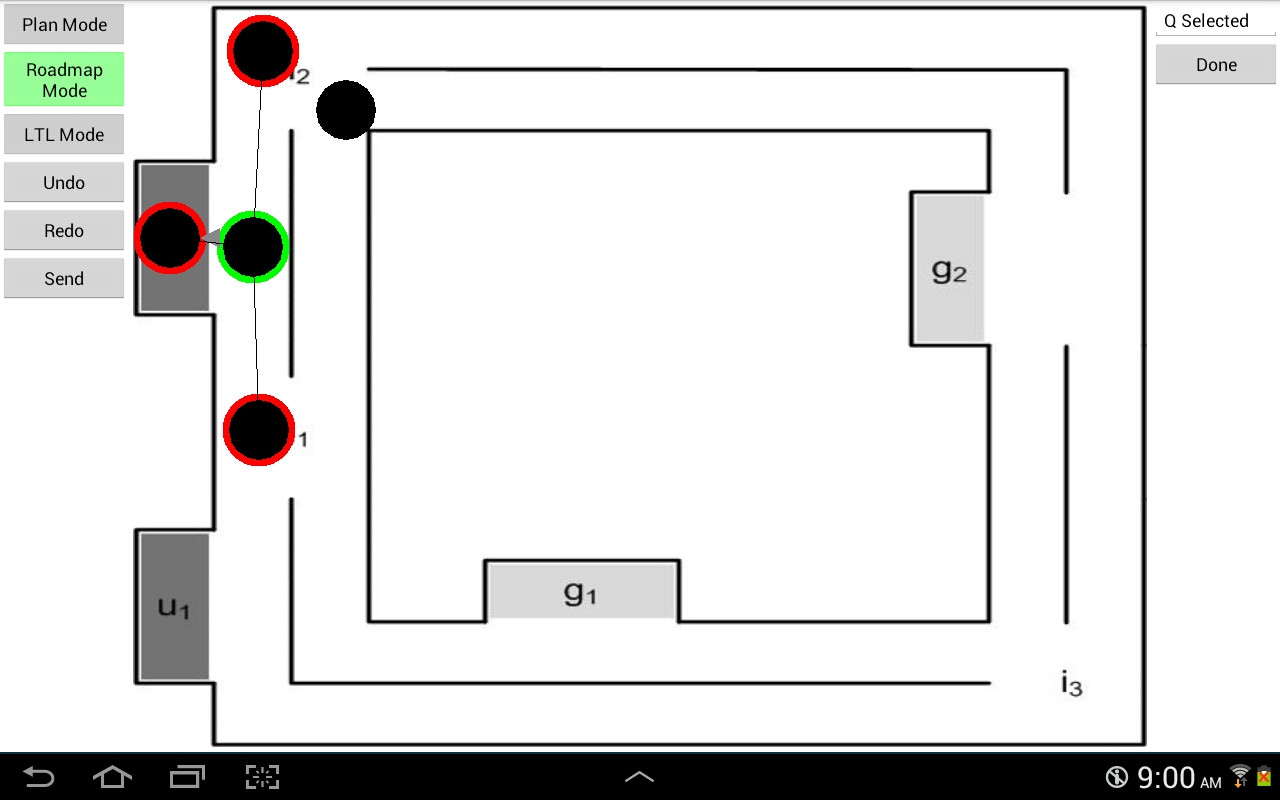}
\end{subfigure}
%\hfill
\begin{subfigure}[b]{0.2\textwidth}
\centering
\includegraphics[width=\textwidth]{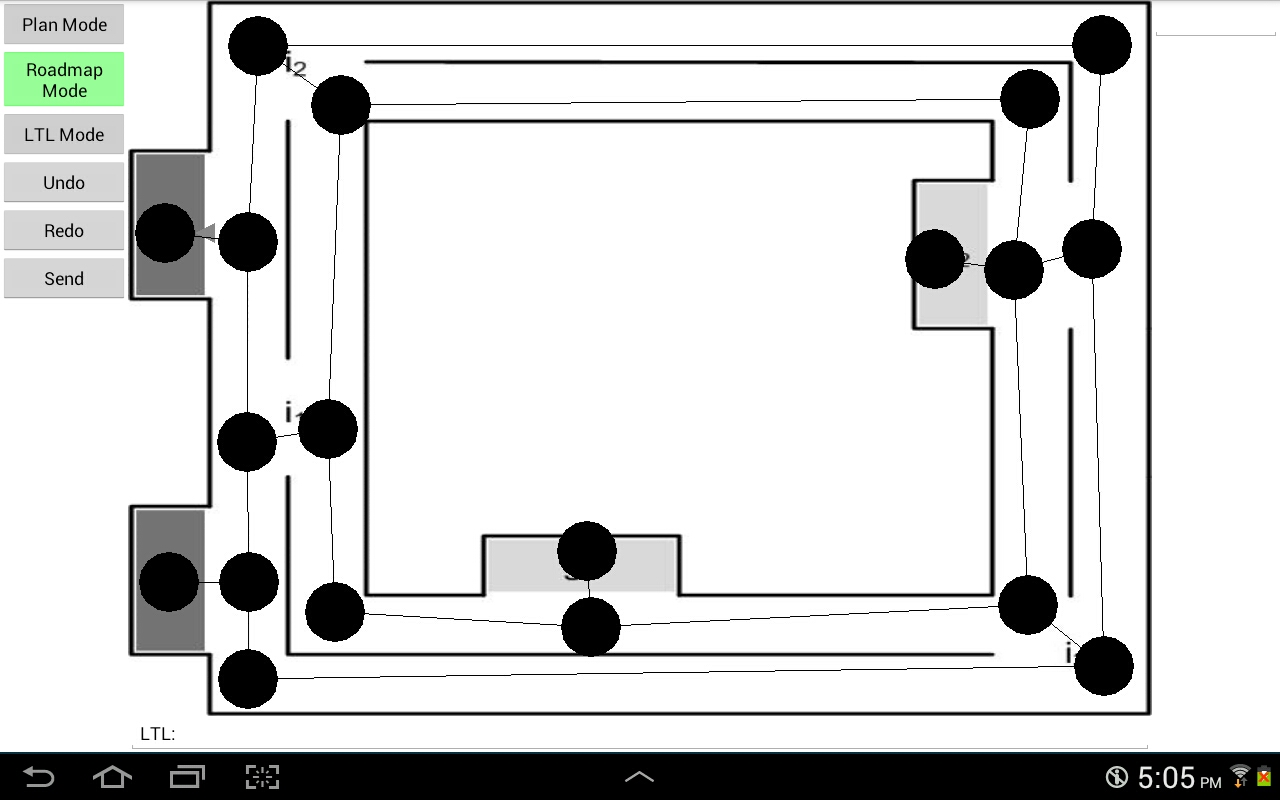}
\end{subfigure}
\caption{The procedure of manually creating a roadmap.
Upper left: the user is asked to load a map.
Upper right: when a node is selected, it will be colored green.
Lower left: the red nodes are denoted as neighbors of the green node.
Lower right: a complete roadmap.}
\label{fig:manual_procedure}
\end{figure}
}
\begin{itemize}
\item Sketching Mode (Fig. \ref{fig:skteched_path}): Create nodes; Move nodes; Draw a path from one node to another; Calculate the most suitable path according to the user drawing; Clear current drawing and planning path.
\item Roadmap Mode (Fig. \ref{fig:manual_procedure}): Create nodes; Add or remove undirected edges between nodes; Automatically save once switching to another mode.
\item LTL Mode (Fig. \ref{fig:default_ltl_symbol}): Create nodes; Add or remove edges with LTL attributes; Edit LTL attributes.
\end{itemize}

After loading the map, the interface enters the roadmap mode.
%\shl{A roadmap is an editable and storable transition system.}
\yhl{A roadmap, which is editable and serializable through the interface, represents the TS.}
For example, the roadmap in Fig. \ref{fig:manual_procedure} is a topological graph which represents the workspace of the robot.
The roadmap should be stored locally as roadmap data.
If the roadmap data exists, it will be loaded and then the sketching mode will be entered; otherwise, the interface will enter the roadmap mode and automatically create an empty roadmap data for editing.
When a user is done editing a roadmap, the interface will switch to sketching mode.
The last step is to create an LTL specification.
However, there is no restriction for the accessing order of each mode.
A user can access any mode at any time.

\section{Extended-LTLvis}
%In this section, an extended LTLvis (E-LTLvis) will be introduced.
E-LTLvis enables several drawing features and different interface layouts from the original LTLvis \cite{SrinivasKKKF2013}.

\subsection{Load Map and Create Roadmap}
Roadmaps can be automatically generated using grid decomposition or a polyhedral decomposition of the environment \cite{LaValle2006}.
In our interface, we require user to manually create their own roadmap.
First, the interface requires the user to load the roadmap image when the interface starts (Fig. \ref{fig:manual_procedure} top left).
After the image is loaded, the interface will search the corresponding roadmap file (.spc) which stores roadmap data.
If it exists, the data is loaded.
If it does not exist, the interface will switch to roadmap mode and automatically create an empty roadmap data to allow the user to edit.
When finishing editing the map, the roadmap file will be created to store these nodes and edges locally.
Next time, when the same map image is selected, this roadmap file will be loaded automatically.
We also provide a video demo in \cite{eLTLvisWiki} to show in more detail the procedure for creating a roadmap.
Figure \ref{fig:manual_procedure} contains some screen shots of this demo.

\subsection{Sketch Path}
Sketch mode allows users to customize the path between any two nodes.
In sketching mode, users can add a node by long pressing on the screen.
When customizing the path, you can first select the starting node, drag the path along the map, and end the path at another node.
This path is denoted as \textbf{\textit{user sketched path} $p^u$}.
Then, we find the node in the environment closest to the first node of $p^u$, and denote it as $q_{start}$.
Also, we find the node in the environment closest to the last node of $p^u$, and denote it as $q_{end}$.
Because the user sketched path may be drawn by curves which consist of too many nodes, to reduce the computation workload, the path is sampled by distance $d_m$ and angle $\theta_m$ into a list of (blue in the Fig. \ref{fig:blacknodesandedges}) nodes $(n1, n2, \dots)$.
After appending $q_{start}$ to the beginning of the list and $q_{end}$ to the end of the list, we get a new list of nodes.
This list of nodes is denoted as \textbf{\textit{sampled user sketched path $p^{0}$}}.
For example, in Fig. 3, the green curve is the user sketched path.

%\todo[inline]{TODO: Add Figure 3.}

\ifthenelse {\boolean{TECHREP}}
{
\begin{figure}
\centering
\includegraphics[width=0.24\textwidth]{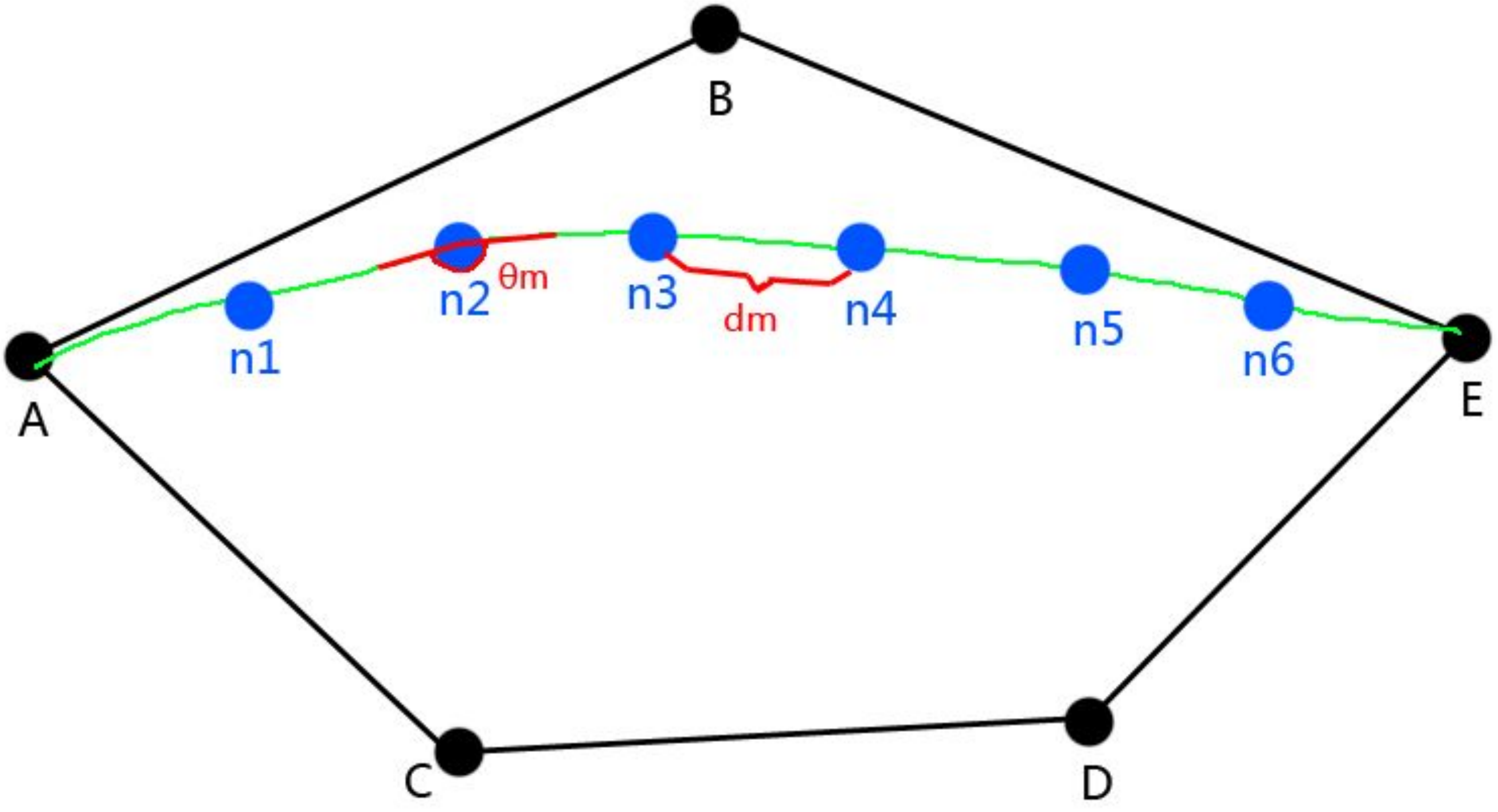}
\caption{Black nodes and edges: the roadmap of a simple environment. Green path: $p^u$. Blue nodes: $p^0$}
\label{fig:blacknodesandedges}
\end{figure}
}
{
\begin{figure}
\centering
\includegraphics[width=0.2\textwidth]{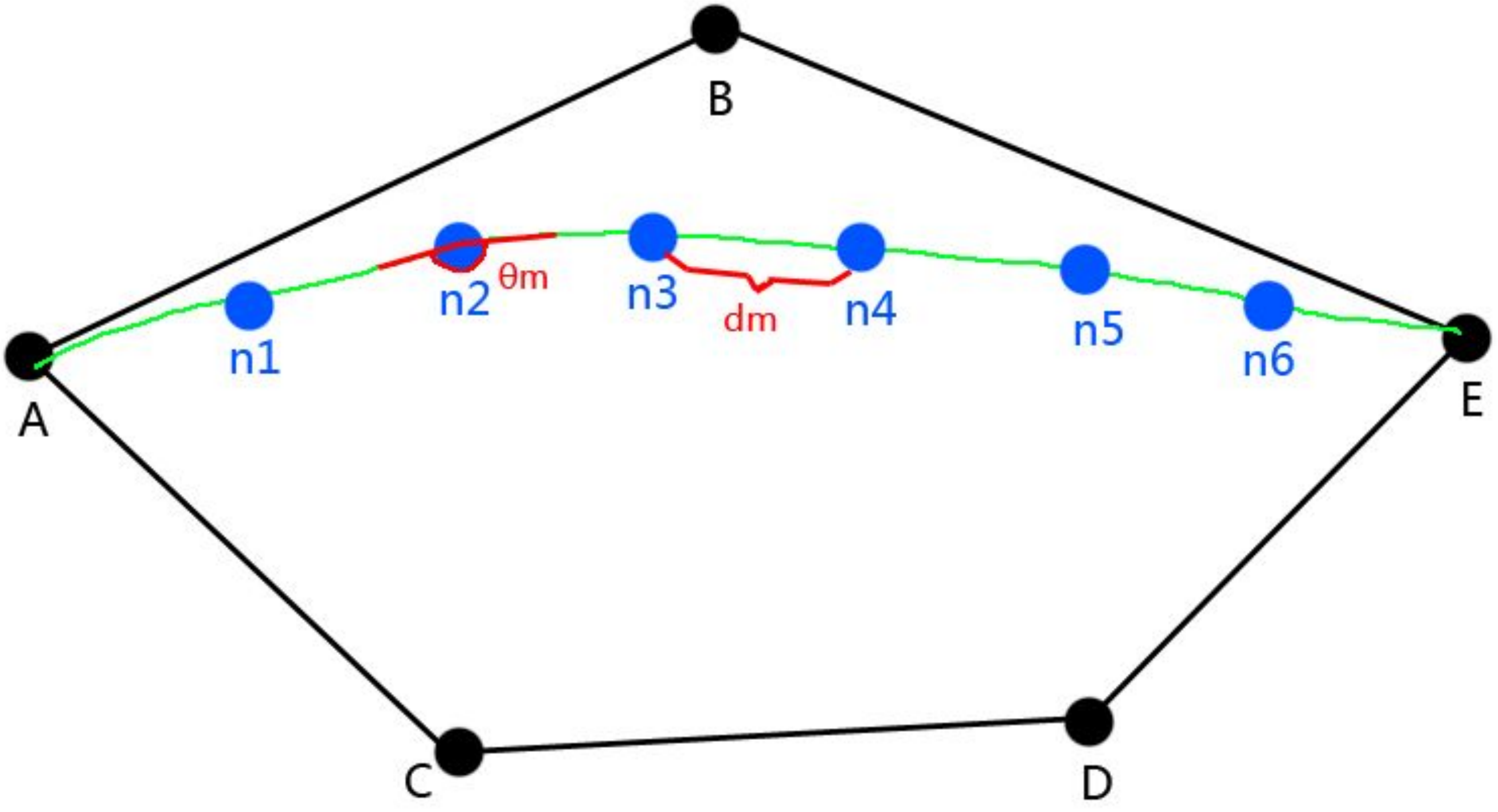}
\caption{Black nodes and edges: the roadmap of a simple environment. Green path: $p^u$. Blue nodes: $p^0$}
\label{fig:blacknodesandedges}
\end{figure}
}

Then, the touch up event will be triggered and the computed best matching path will be displayed.
Since $q^u$ may stretch to areas undefined in the roadmap, this path may not be the same as $q^u$ (Fig. \ref{fig:skteched_path}).
As we need to compare the similarity of two paths, the best approach is to calculate the volume between two paths.
However, this approach has heavy workload.
Thus, we define a new heuristic, CWPD, to compare two paths.

\begin{defn} (CWPD) Component-Wise Path Distance is a
distance between two paths $p^0 = (n^0_0, n^0_1, \dots, n^0_{N-1})$ and $p^x = (n^x_0, n^x_1, \dots, n^x_{N-1})$.
It is defined as:
\begin{equation}
\label{eq1}
CWPD(p^0,p^x) = \sum_{i=0}^{N-1}dist(n^0_i,e_{(n^x_j,n^x_i)})),
\end{equation}
%where $N = size(p^0), n^0_i \in p^0, n^x_i, n^x_j \in p^x, \text{ and } n^x_j$ is previous node which differs from $n^x_i$. If $n^x_i$ is the first node, $n^x_j$ equals to $n^x_i$. 
where $N$ $=$ $size(p^0)$, $n^0_i$ $\in$ $p^0$, $n^x_i$, $n^x_j$ $\in$ $p^x$, and $n^x_j$ is previous node which differs from $n^x_i$. If $n^x_i$ is the first node, $n^x_j$ equals to $n^x_i$. 
%\[CWPD(p^0,p^x) = \sum_{i=0}^{N-1}distance(n^0_i,edge_{(n^x_j,n^x_i)}))\],
\end{defn}

We remark that a path can have repetition of nodes. 
We use distance to line segment $(e_{(n^x_j,n^x_i)})$ instead of line to avoid
the situation where $n^0_i$ is very far from $e_{(n^x_j,n^x_i)}$ but close to the $line(n^x_j,n^x_i)$.
From Eq.~(\ref{eq1}), we can also derive the following equation.
Let $CWPD\Big((n^0_0,$ $n^0_1,$ $\dots,$ $n^0_{N-1}),$ $(n^x_0, n^x_1, \dots, n^x_{N-1})\Big)$ denote $A$ and $CWPD\Big((n^0_0, n^0_1, \dots, n^0_{N-2}),(n^x_0, n^x_1, \dots, n^x_{N-2})\Big) + dist(N^0_{N-1}, e_{(n^x_{j-1},n^x_{N-1})})$ denote $B$. Then,

\begin{equation}
\label{eq2}
\begin{split}
A & = \sum_{i=1}^{N-1}dist(n^0_i, e_{(n^x_j, n^x_i)}) \\
 & = \Big(\sum_{i=1}^{N-2}dist(n^0_i, e_{(n^x_j, n^x_i)})\Big) + dist(n^0_{N-1}, e_{(n^x_{j-1}, n^x_{N-1})}) \\
 & = B
\end{split}
\end{equation}

Then, we definite the best match path in order to compare it in terms of distance.

\begin{defn} (BMP) Best Matching Path $p_{bmp}$ is a feasible path on the transition system TS with the same starting $q_{init}$ and ending position $q_{end}$ as $P^0$.
It also has the properties: $length(p_{bmp})$ $=$ $length(p^0)$; $p_{bmp}$ can be cyclic on TS; The \textbf{\textit{component-wise path distance}} between $p_{bmp}$ and $p^0$ should be minimal.
\end{defn}

We can reduce the sample distance $d_m$ and angle $\theta_m$ to increase $N$. Thus, $p^0$ can always have more nodes than $p_{bmp}$ so that the size of $p_{bmp}$ can be extended to $N$ by adding copies of nodes in between.
For the example in Fig. \ref{fig:cwpds}, some possible BMP candidates are listed in Table \ref{tb1} for the candidate path set $(p^1, p^2, p^3)$:

\begin{table}[h!]
\centering
\begin{tabular}{|c|c|c|c|c|c|c|c|c|}
\hline
$p^1$	& A  & B  & B  & B  & B  & E  & E  & E \\
\hline
$p^2$	& A  & C  & C  & D  & D  & E  & E  & E \\
\hline
$p^3$	& A  & B  & B  & B  & B  & B  & E  & E \\
\hline
$p^0$	& A  & n1 & n2 & n3 & n4 & n5 & n6 & E \\	
\hline 
\end{tabular}
\vspace{5pt}
\caption{path $p^0$ and its possible BMP candidates. As $p^0$ has more nodes than $p^x$, we can extend the path (A, B, C) to path (A, B, B, \dots, C) or (A, B, C, \dots, C) to make their number of nodes equal to $N$. To achieve the minimum CWPD, we need to compare the CWPDs (shown in Fig. \ref{fig:cwpds}) between $p^0$ and each $p^x$. In this example, the path $p^3$ minimizes the CWPD.}
\label{tb1}
\end{table}

%\todo[inline]{TODO: add Figure 4, here.}

\ifthenelse {\boolean{TECHREP}}
{
\begin{figure}
\centering
\begin{subfigure}[b]{0.24\textwidth}
\centering
\includegraphics[width=\textwidth]{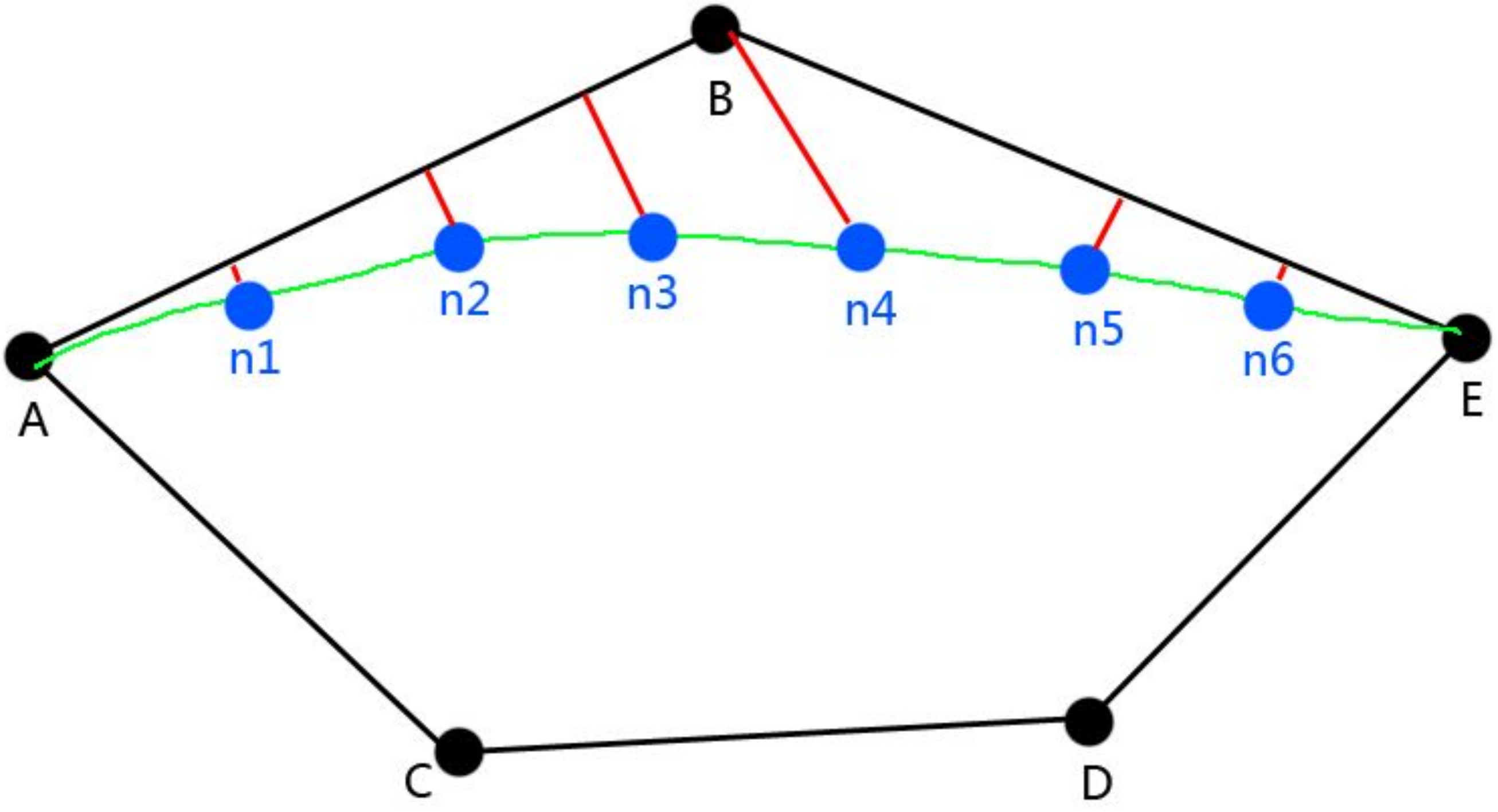}
\end{subfigure}
%\hfill
\begin{subfigure}[b]{0.24\textwidth}
\centering
\includegraphics[width=\textwidth]{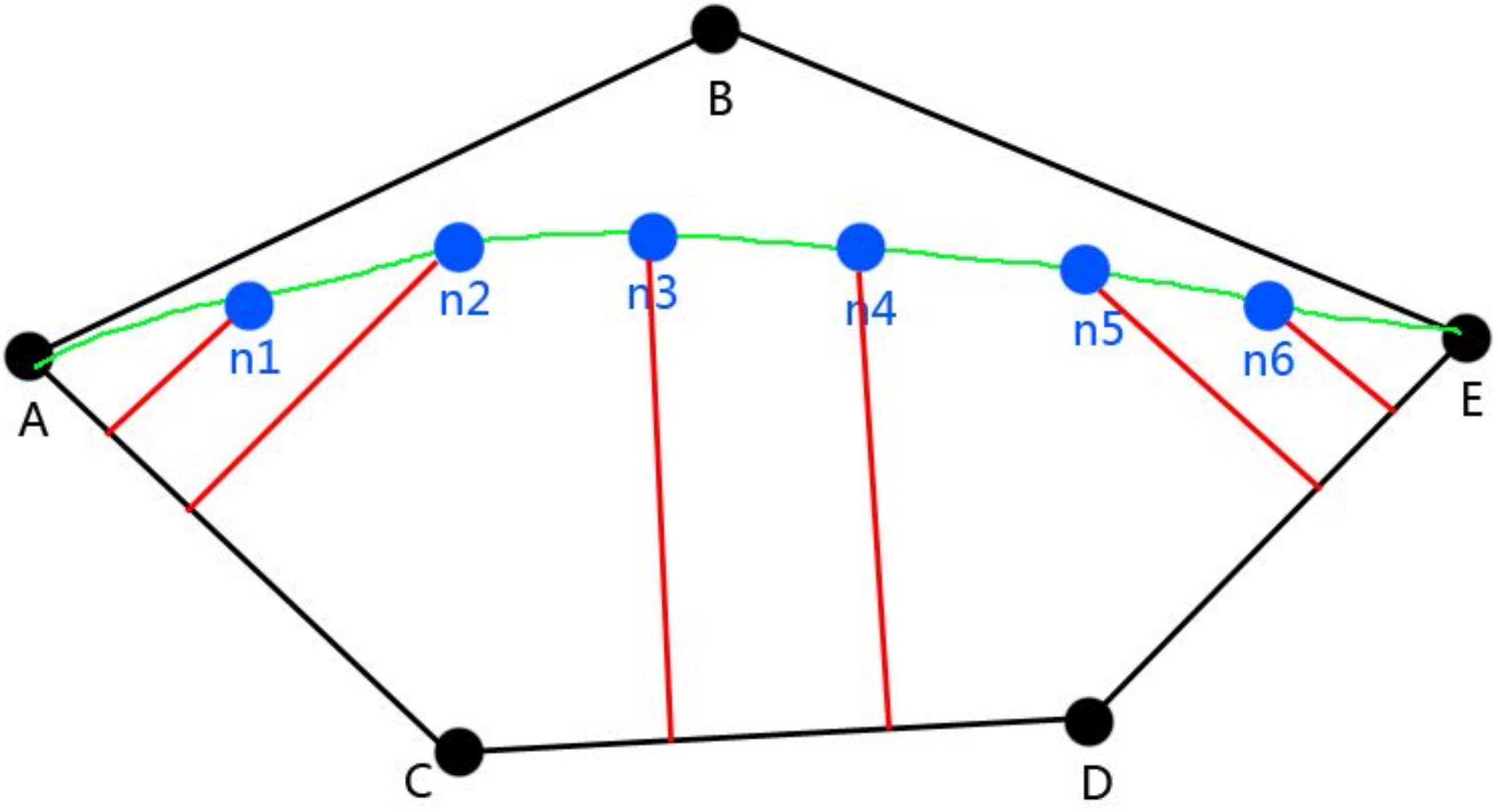}
\end{subfigure}
\vskip\baselineskip
\centering
\begin{subfigure}[b]{0.24\textwidth}
\centering
\includegraphics[width=\textwidth]{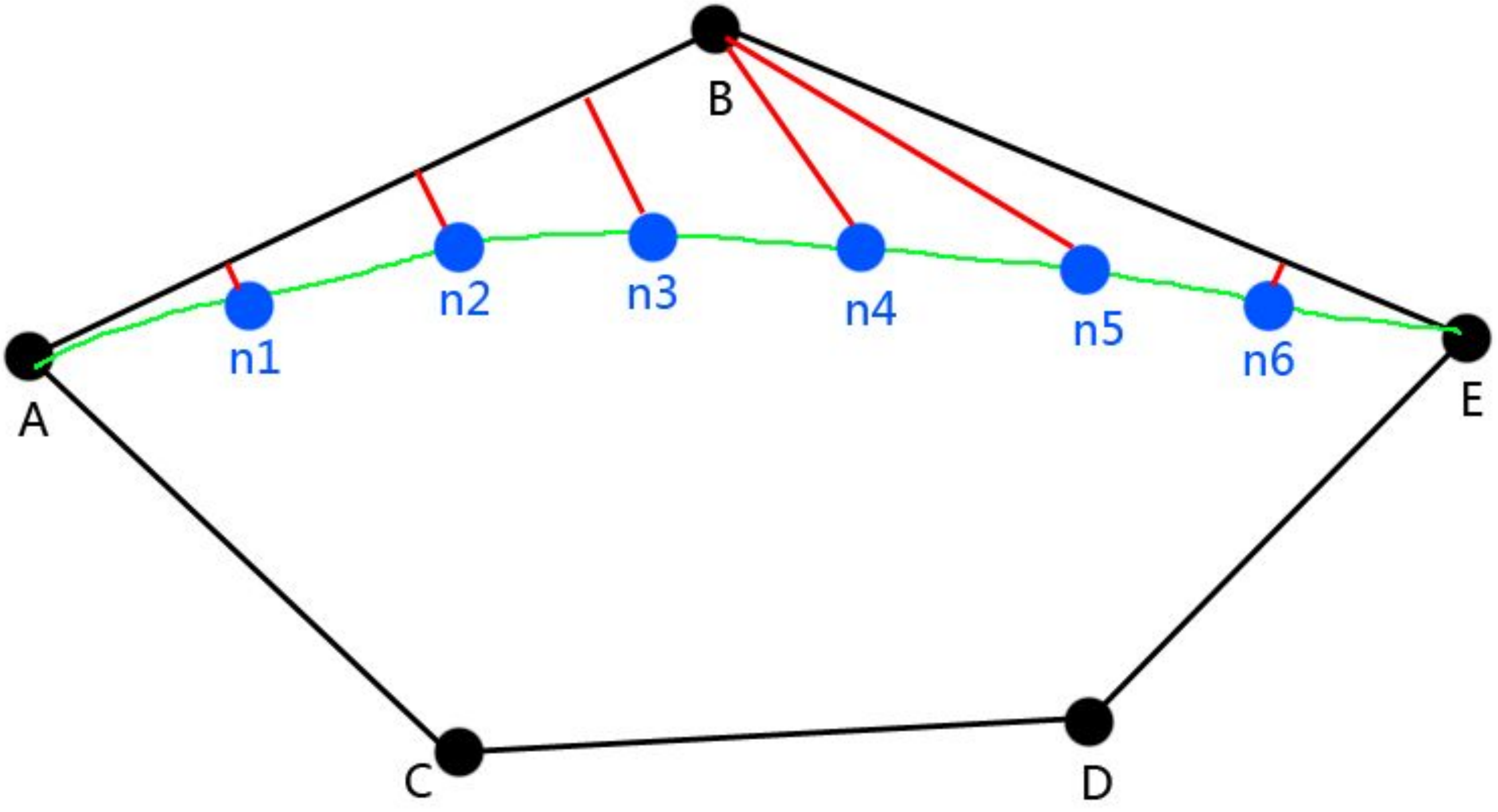}
\end{subfigure}
\caption{The CWPDs of $p^1$, $p^2$, $p^3$}
\label{fig:cwpds}
\end{figure}
}
{
\begin{figure}
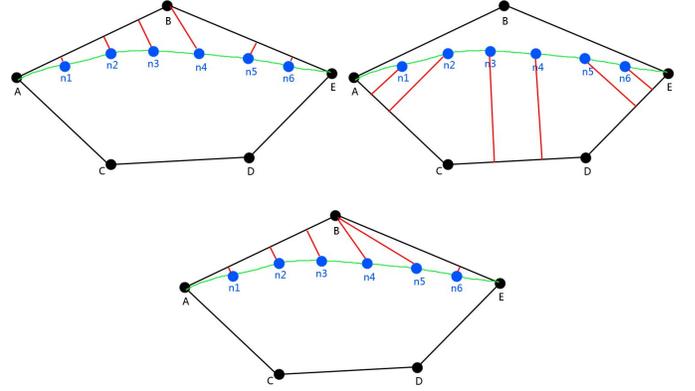

\centering
\begin{subfigure}[b]{0.2\textwidth}
\centering
\includegraphics[width=\textwidth]{figure6_1}
\end{subfigure}
%\hfill
\begin{subfigure}[b]{0.2\textwidth}
\centering
\includegraphics[width=\textwidth]{figure6_2}
\end{subfigure}
\vskip\baselineskip
\centering
\begin{subfigure}[b]{0.2\textwidth}
\centering
\includegraphics[width=\textwidth]{figure6_3}
\end{subfigure}
\caption{The CWPDs of $p^1$, $p^2$, $p^3$}
\label{fig:cwpds}
\end{figure}
}

As the number of candidate in the worst case is $N^M$, where M is the number of nodes in TS, it is impractical to list all of them before searching the minimum CWPD.
Instead, we create a matrix to store a BMP ending at $n^x_i$ for each node in the roadmap.
When looping through each node in $p^0$, the path stored in the matrix will be updated.
The pseudo code of this greedy algorithm is provided in Alg. \ref{alg:findbmp}.

\begin{algorithm}
\caption{\text{\sc FindBMP}$(p^0, TS)$}
\label{alg:findbmp}
\begin{algorithmic}[1]
\Require a path $p^0$ and $TS$ $:=$ $(Q_{TS}$, $q_{init}$, $\delta_{TS}$, $\Pi$, $h$, $w_{TS}$)
\Ensure a path $p_{bmp}$

\State $M \gets |Q_{TS}|$ \Comment the number of elements in $Q_{TS}$
\State $N \gets |p^0|$ \Comment the number of nodes in $p^0$
\State $cwpd[:,:] \gets \infty$ \Comment for $N \times M$ matrix
\State $bmp[:,:] \gets \emptyset$ \Comment for $N \times M$ matrix
\State $\tupleof{start, end} \gets \tupleof{index(p^0[1]),index(p^0[N])}$
\State \Comment get indices for $start$ and $end$ from nodes in $Q_{TS}$
\State $\tupleof{cwpd[1,start], bmp[1,start]} \gets \tupleof{0,\{p^0[1]\}}$
\For {$i = 2$ to $N$} \label{alg:findbmp:for1}
\For {$j = 1$ to $M$} \label{alg:findbmp:for2}
\State $\text{\sc Update}(cwpd, bmp, i, j, p^0, TS)$
\EndFor
\EndFor
\State $p_{bmp} \gets bmp[N,end]$
\State \Return $p_{bmp}$
\end{algorithmic}
\end{algorithm}

Because the user sketched path may contain cycles intentionally, standard shortest path algorithms \cite{CormentLRS2009} cannot be used.
Algorithm \ref{alg:findbmp} solves the problem also with cycles on the graph.
It takes $p^0$ and $TS$ as input.
It proceeds sequentially through all nodes in $p^0$ (line \ref{alg:findbmp:for1}).
In each iteration of this outer loop, it calculates M BMPs for each $q_j$ (line \ref{alg:findbmp:for2}) according to current user input path $(n^0_0,$ $n^0_1,$ $\dots,$ $n^0_{i-1})$.
These BMPs start from $q_{start}$ and end at $q_j$.

\begin{algorithm}
\caption{\text{\sc Update}($cwpd, bmp, i, j, p^0, TS)$}
\label{alg:update}
%\algsetup{indent=1em}
\begin{algorithmic}[1]
\Require two matrix $cwpd$ and $bmp$, two variables $i$ and $j$, a path $p^0$ and $TS := (Q_{TS}, q_{init}, \delta_{TS}, \Pi, h, w_{TS})$
\Ensure

\If {$bmp[i-1, j] \neq \emptyset$} \Comment prev. bmp ending at this node
\State $q_j \gets index^{-1}(j, Q_{TS})$ \Comment returns a node of $Q_{TS}$
\State $n^0_i \gets index^{-1}(i, p^0)$ \Comment returns a node of $p^0$
\State $e_{prev} \gets \text{\sc GetLastEdge}(bmp[i-1,j])$ \label{alg:update:getlastedge}
\State $cwpd_{candi} \gets cwpd[i-1,j] +$$dist(n^0_i, e_{self})$ \Comment \text{Eq. (\ref{eq2})}
\If {$cwpd_{candi} < cwpd[i,j]$}
\State $cwpd[i,j] \gets cwpd_{candi}$
\State $bmp[i,j] \gets bmp[i-1,j] + q_j$
\State \Comment concatenates $q_j$ to the end of $bmp[i-1,j]$
\EndIf
\For {$q_k$ in $Neighbors(q_j)$}
\State $e_{curr} \gets \tupleof{q_j, q_k}$
\State $k \gets index(q_k)$ \Comment index of nodes in $Q_{TS}$
\If {$e_{curr} \neq e_{prev}$}
\State $cwpd_{candi} \gets cwpd[i-1,k] + dist(n^0_i, e_{curr})$
\If {$cwpd_{candi} < cwpd[i,k]$}
\State $cwpd[i,k] \gets cwpd_{candi}$
\State $bmp[i,k] \gets bmp[i-1,k] + q_k$
\EndIf
\EndIf
\EndFor
\EndIf
\end{algorithmic}
\begin{algorithmic}
\\\hrulefill
{\scriptsize
\begin{itemize}
\item At line \ref{alg:update:getlastedge}, \text{\sc GetLastEdge}() returns the last edge of a given path or an edge with the same two nodes if there is no last edge e.g., \text{\sc GetLastEdge}([ABCDE]) returns [AB] and \text{\sc GetLastEdge}([A]) returns [AA].
\end{itemize}
}
\end{algorithmic}
\end{algorithm}

Algorithm 2 calculates the new CWPD and BMP by utilizing the results from the previous BMPs and CWPDs using \text{Eq. (\ref{eq2}).}
For each node $q_j$ in $Q_{TS}$, it first checks if $q_j$'s previous BMP $bmp[i \text{ -- } 1, j]$ for $(p^0[1], \dots, p^0[i \text{ -- } 1])$ exists.
If it exists, it calculates the distance between $p^0[i]$ and the last edge of the path $(bmp[i \text{ -- } 1, j], q_j)$.
Then, it stores the result in $bmp[i, j]$ and $cwpd[i, j]$ if the new $cwpd[i, j]$ is smaller than the existing value.
Then, it repeats the process for all paths ($bmp[i \text{ -- } 1, j], q_k)$, where $q_k \in Neighbors(q_j)$.
Note that we can get $q_j$'s previous BMP and CWPD directly from $bmp[i \text{ -- } 1, j]$ and $cwpd[i \text{ -- } 1, j]$, respectively, without recomputing the results.
The process will repeat at most M times; thus, the run time of Alg. \ref{alg:update} is $O(M)$.
In each step, the minimum CWPD ending at each node in TS will be stored. Thus, this algorithm finds the BMP with the minimum CWPD eventually.
The step by step run of Alg. \ref{alg:findbmp} over the example of Table \ref{tb1} can be found in \cite{WeiThetis2016}.
The algorithm only creates two global matrices of size NM.
Thus, the space complexity of this algorithm is $O(NM)$ and the runtime complexity is $O(NMM)$.
Hence, this algorithm can be implemented on a mobile device.
After applying the algorithm to the scenario in Fig. \ref{fig:manual_procedure}, we can get the result in Fig. \ref{fig:skteched_path}.

%\todo[inline]{TODO: add Figure \ref{fig:skteched_path} below.}
\ifthenelse {\boolean{TECHREP}}
{
\begin{figure}
\centering
\includegraphics[width=0.40\textwidth]{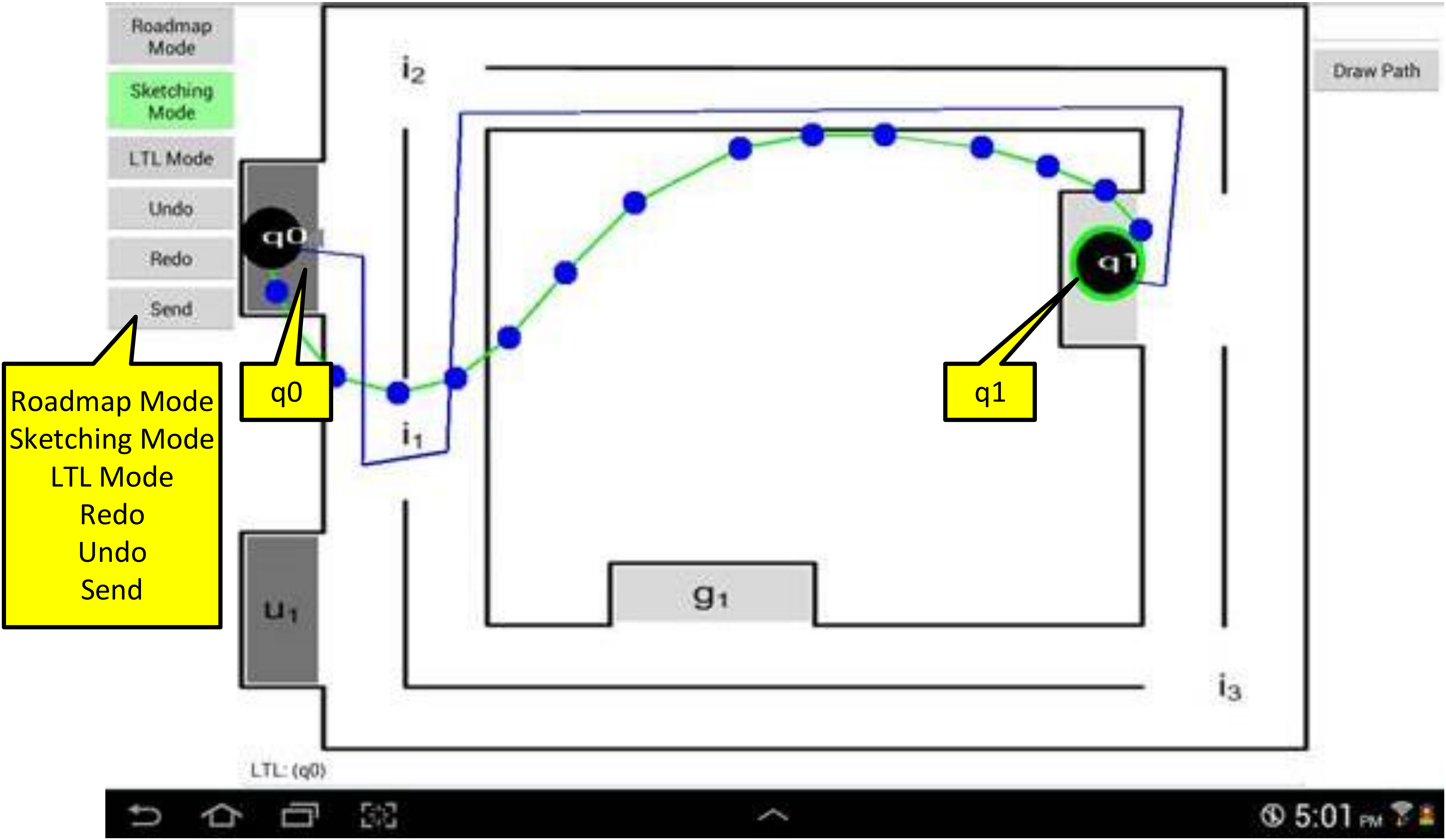}
\caption{The user sketched path (arc with dots) and its BMP (solid line in the middle of lane). Since the user sketches in areas undefined in the roadmap, the resulting BMP is much different from the sketched path. The textboxes are added to improve the readability due to the size of the screenshot.}
\label{fig:skteched_path}
\end{figure}
}
{
\begin{figure}
\centering
\includegraphics[width=0.32\textwidth]{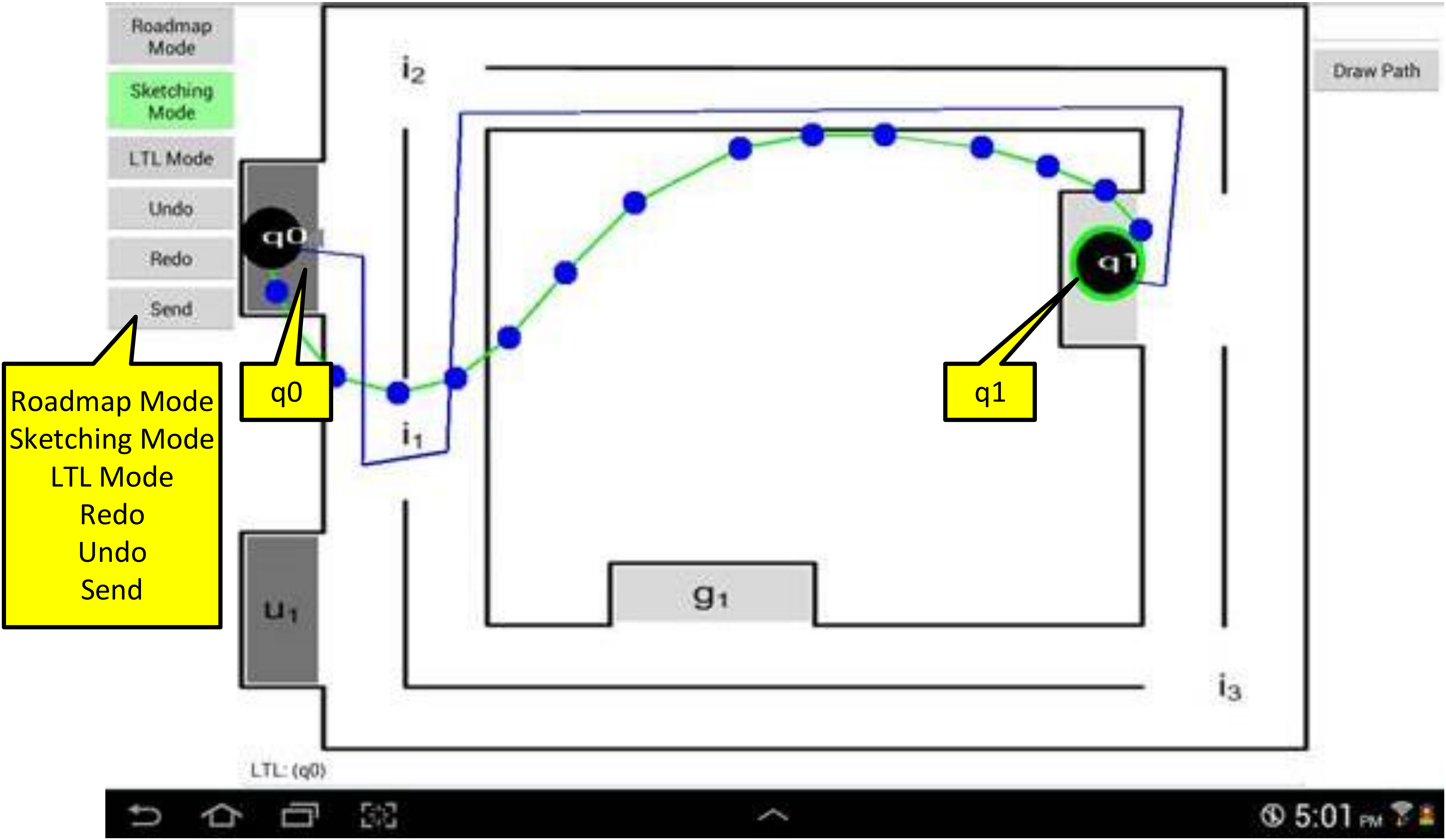}
\caption{The user sketched path (arc with dots) and its BMP (solid line in the middle of lane). Since the user sketches in areas undefined in the roadmap, the resulting BMP is much different from the sketched path. The textboxes are added to improve the readability due to the size of the screenshot.}
\label{fig:skteched_path}
\end{figure}
}
Usually, users may need to specify the paths between multiple pairs of nodes.
Our algorithm will generate multiple best matching paths for all user sketched paths.
This set of best matching paths is called the \textbf{\textit{preferred path set}}.

\subsection{Edit Specifications}
After a path is customized in the Sketching mode, there should be a default LTL specification displayed in the LTL Mode (for an example see Fig. \ref{fig:default_ltl_symbol}).
%\todo[inline]{TODO: add Figure 6 below.}
\ifthenelse {\boolean{TECHREP}}
{
\begin{figure}
\centering
\includegraphics[width=0.38\textwidth]{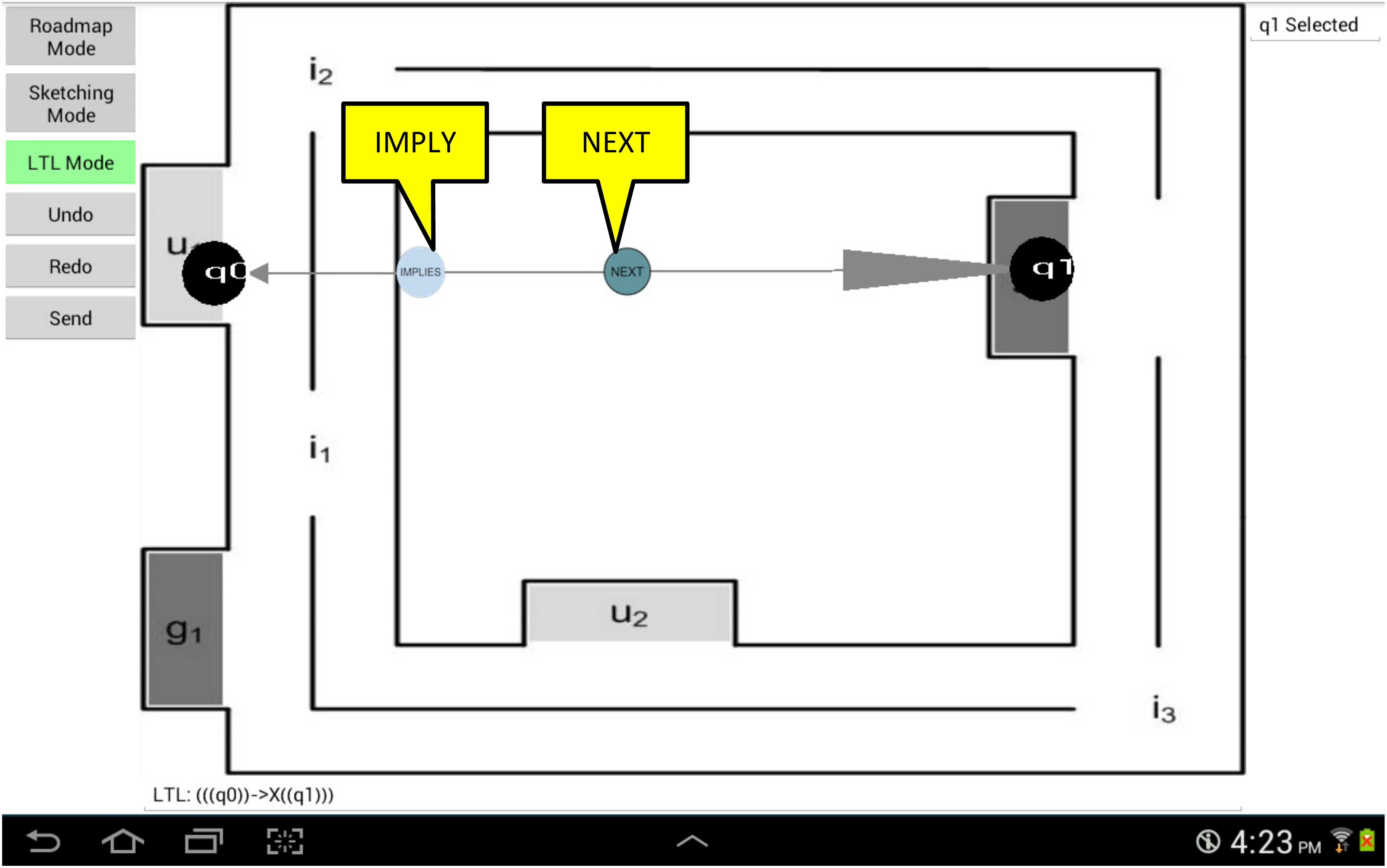}
\caption{The basic LTL specification that corresponds to the sketched path in Fig. \ref{fig:skteched_path}}
\label{fig:default_ltl_symbol}
\end{figure}
}
{
\begin{figure}
\centering
\includegraphics[width=0.3\textwidth]{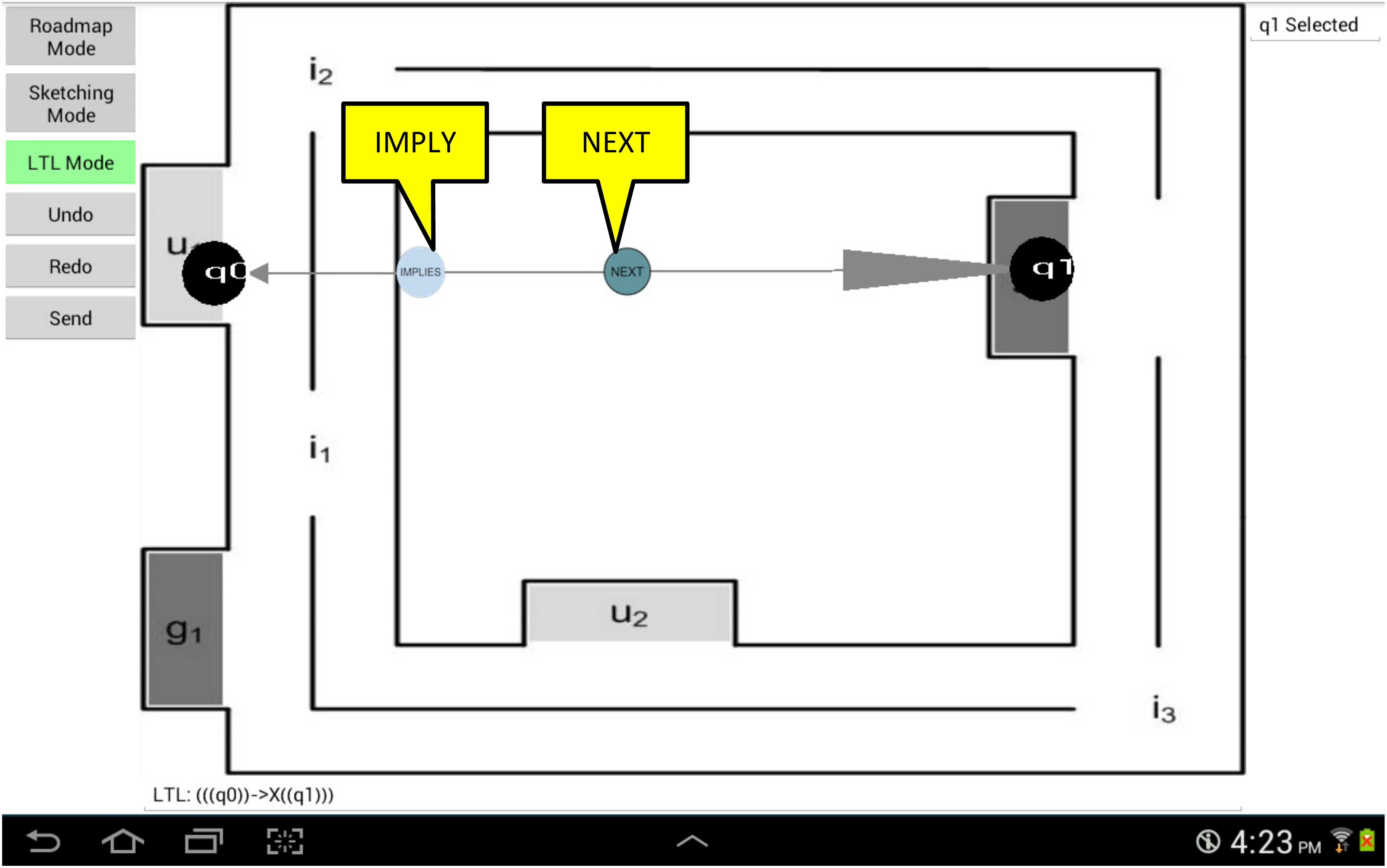}
\caption{The basic LTL specification that corresponds to the sketched path in Fig. \ref{fig:skteched_path}}
\label{fig:default_ltl_symbol}
\end{figure}
}

Users can also skip the Sketching mode to directly edit the LTL specification.
In this mode, the editing gestures are identical to LTLvis \cite{SrinivasKKKF2013}.

\subsection{Send Data}
When all the data is ready, users can send the data to the LTL planner.
The LTL planner used in this work is modified from the RHTL package \cite{UlusoyMB2013}.
By adding path preference logic in the traditional LTL planner, the resulting path generated from the new planner will attempt both the LTL specification and the user input requirement.
%The details of this planner are explained in the next section.

\section{Planning Using-E-LTLvis}
In this section, we will explain an extended planner (Alg. \ref{alg:e_planner}) which is modified from a RHTL package \cite{UlusoyMB2013}.
It takes the product automaton $A$, the local transition system $TS$ and the preferred path set D as inputs.
In this algorithm, $A$ will be a tuple $A: = (\varPsi, q_{init}, \delta, W, F)$.
We denote a preferred path set as $D$, where $D$ $:=$ $\{\pi_{uv}\ |\ \pi_{uv}$ $=$ $(q_u$, $n_{a1}$, $n_{a2}$, $\dots$, $n_{am}$, $q_v$), $q_u$, $q_v$ $\in \varPsi$, $n_{a1}$, $n_{a2}$, $\dots$, $n_{am}\in Q_{TS} \}$.

\begin{algorithm}
\caption{$\text{\sc ExtendedPlanner}(A, TS, D)$}
\label{alg:e_planner}
\begin{algorithmic}[1]
\Require a product automaton $A := (\varPsi, q_{init}, \delta, W, F)$, a local transition system $TS := (Q_{TS}, q_{init}, \delta_{TS}, \Pi, h, w_{TS})$ and a preferred path set $D$ $:=$ $\{\pi_{uv}\ |\ \pi_{uv}$ $=$ $(q_u$, $n_{a1}$, $n_{a2}$, $\dots$, $n_{am}$, $q_v$), $q_u$, $q_v$ $\in \varPsi$, $n_{a1}$, $n_{a2}$, $\dots$, $n_{am}\in Q_{TS} \}$
\Ensure an extended path $\pi_{ltl}$

\State Create an empty list $\pi_{ltl}$
\For {each $\pi_{ij}$ in $D$} \Comment set all the preferred paths to highest priority to be chosen
\State $\tupleof{q_i,q_j} \gets \tupleof{\pi_{ij}[1],\pi_{ij}[|\pi_{ij}|]}$
\If {$\tupleof{q_i,q_j} \in \delta$} \label{alg:e_planner:in_delta}
\State Change the weight $w(q_i,q_j)$ to $\alpha$ \label{alg:e_planner:alpha}
\EndIf
\EndFor
\State Find the shortest path $\pi_{A0}$ with minimum sum of edge weight from $q_{init}$ to $q_{accept}$ in $A$ \Comment based on the modified priorities above \label{alg:e_planner:find_pi_a0}
\For {$k = 1$ to $|\pi_{A0}| - 1$}
\State $\tupleof{q_h,q_m} \gets \tupleof{\pi_{A0}[k],\pi_{A0}[k+1]}$
\State $found \gets \bot$
\For {$\pi_D$ in $D$} \Comment $\pi_D$ is a sequence of nodes in $Q_{TS}$

%\If {$\pi_{D}[1]$=$q_h$$\land$$\pi_D[|\pi_D|]$=$q_m$$\land$$\pi_D$ is valid in $A$} \label{alg:e_planner:valid}
\If {$\pi_{D}[1]$=$q_h \land \pi_D[|\pi_D|]$=$q_m \land$$\text{\sc Valid}(\pi_D,A)$} \label{alg:e_planner:valid}
%\If {$\pi_{D}[1] = q_h$ and \pi_D[|\pi_D|] = q_m$ and $\pi_D$ is valid in $A$} \label{alg:e_planner:valid}
%\If {$\pi_{D}[1] = q_h$ and $\pi_D[|\pi_D|] = q_m$ and $\pi_D$ is valid in $A$} \label{alg:e_planner:valid}
\State Append $\pi_D$ to $\pi_{ltl}$ \label{alg:e_planner:append1}
\State $found \gets \top$
\EndIf
\EndFor
\EndFor
\If {$\neg found$}
\State Find the shortest path $\pi'_D$ from $q_h$ to $q_m$ in $TS$
\State Append $\pi'_D$ to $\pi_{ltl}$ \label{alg:e_planner:append2}
\EndIf
\State Append $\pi_{A0}[|\pi_{A0}|]$ to $\pi_{ltl}$ \Comment this is for $q_{accept}$ \label{alg:e_planner:append3}
\State \Return $\pi_{ltl}$
\end{algorithmic}

\begin{algorithmic}
\\\hrulefill
{\scriptsize
\begin{itemize}
\item At line \ref{alg:e_planner:alpha}, $\alpha$ is infinitesimal and $\alpha \in \mathbb{R}_+$. It is much smaller than the smallest weight in $W$.
\item At line \ref{alg:e_planner:valid}, $\text{\sc Valid}(\pi_D,A)$ means that this path $\pi_D$ never visits any avoiding states in $A$.
\item At line \ref{alg:e_planner:append1}, \ref{alg:e_planner:append2}, \ref{alg:e_planner:append3}, each Append operation to $\pi_{ltl}$ adds the element to the tail of the list.
\end{itemize}
}
\end{algorithmic}
\end{algorithm}

Algorithm \ref{alg:e_planner} works as follows.
Assume $\pi_{ij} \in D$, the algorithm first checks if there is transition $(q_i, q_j) \in \delta$ at line \ref{alg:e_planner:in_delta}. 
If such transition exists, it changes its weight to $\alpha$.
Here, $\alpha \in \mathbb{R}_+$ denotes an infinitesimal value.
This can increase the priority of the preferred path set when calculating the shortest path $\pi_{A0}$ from $q_{init}$ to $q_{accept}$ in line \ref{alg:e_planner:find_pi_a0}.
After finding $\pi_{A0}$, we need to replace each transition $(q_i, q_j) \in \pi_{A0}$ with a corresponding transition from either the preferred path set $D$ or the transition system $TS$.
As the preferred path set $D$ has higher priority, if $\pi_D$ exists in the preferred set, we add it to the path $\pi_{ltl}$.
Otherwise, we find a shortest alternative $\pi'_D$ in $TS$ and add it to $\pi_{ltl}$.
After $q_{accept}$ is visited, $\pi_{ltl}$ is completed.
\ifthenelse {\boolean{TECHREP}}
{
For detail, see \cite{WeiThetis2016}.
}
{
For detail, see \cite{WeiKGTech2016} and \cite{WeiThetis2016}.
}

\section{Experiments}

%\todo[inline]{TODO: add Figure 7 below.}
\ifthenelse {\boolean{TECHREP}}
{
\begin{figure}
\centering
\begin{subfigure}[b]{0.22\textwidth}
%\centering
\includegraphics[width=\textwidth]{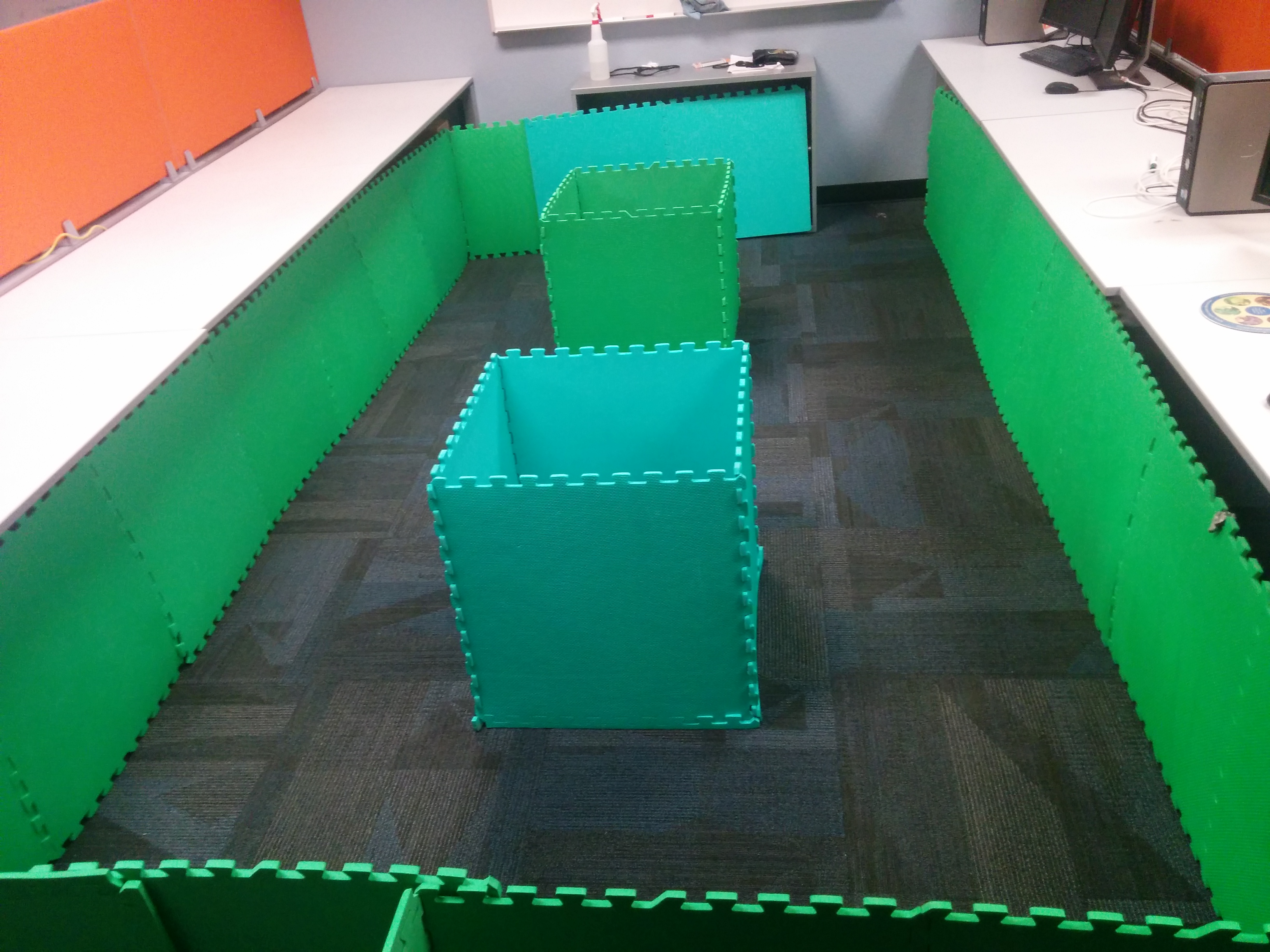}
\end{subfigure}
%\hfill
\begin{subfigure}[b]{0.245\textwidth}
%\centering	
\includegraphics[width=\textwidth]{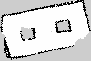}
\end{subfigure}
\caption{The experiment environment and its scanned map}
\label{fig:experiment_environment}
\end{figure}
}
{
\begin{figure}
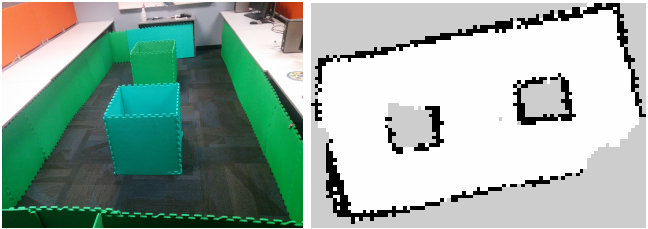

\centering
\begin{subfigure}[b]{0.14\textwidth}
%\centering
\includegraphics[width=\textwidth]{figure13_1}
\end{subfigure}
%\hfill
\begin{subfigure}[b]{0.1583\textwidth}
%\centering	
\includegraphics[width=\textwidth]{figure13_2}
\end{subfigure}
\caption{The experiment environment and its scanned map}
\label{fig:experiment_environment}
\end{figure}
}

In this section, we are going to test our interface and planner on a real robot - TurtleBot.
TurtleBot is a Robot Operating System (ROS) based project.
It contains two major hardware devices: Kinect and iRobot base.
The TurtleBot project also contains many useful packages.
For example, 
\texttt{\footnotesize turtlebot\_navigation} is one of the most popular packages used to localize the robot by itself.
Then, we use \texttt{\footnotesize turtlebot\_rivz} to visualize the environment.
The final goal of this experiment is to use the proposed interface to send an LTL specification and a preferred path set to the planner.
The planner should generate a path plan and order the TurtleBot to execute the plan.
The real environment ($185cm \times 430cm$) and its scanned map are shown in Fig. \ref{fig:experiment_environment}.

%We have run two experiments.
We performed two experiments 10 times each. We measured the time needed to compute BMPs from $p^0$. The number of nodes in $Q_{TS}$ for both scenarios was 6 and the number of nodes in $q^0$ in average were 11.3 and 29, respectively.
%In the first experiment, we provided an LTL specification, a sketched path and its trajectory from the TurtleBot (shown in Fig. \ref{fig:ltl_spec1}).
Figure \ref{fig:ltl_spec1} shows the first experiment.
%The task of the TurtleBot is to execute the specification $(q0 \rightarrow \mathbf{X}q1) \land (q0 \land \mathbf{F}q2) \land (q0 \rightarrow \mathbf{X} \neg q2)$.
The task of the TurtleBot is to execute the specification $(q0 \rightarrow \mathbf{X}q1) \land (q0 \land \mathbf{F}q2)$  (see \cite{FainekosGKP2009,SmithTBR2010} for a description of the temporal logic operators).
%In natural language, it means ``the TurtleBot is required to start from q0 and head for q1 while avoiding q2 before q1 is reached.
%Then it will reach q2 eventually''.
In natural language, it means ``the TurtleBot is required to start from q0 and head for q1.
Then it will reach q2 eventually''.
It took 1.7 milliseconds in average, having M=6 and N=11.3.
%We denote the time for the interface to get the preferred path set as computational time $T_f$.
%Then the $T_f$ for the first experiment is 3.6 milliseconds with M equaling 10 and N equaling 6.
%\todo[inline]{TODO: add Figure 8 below.}

\ifthenelse {\boolean{TECHREP}}
{
\begin{figure}
\centering
\begin{subfigure}[b]{0.3\textwidth}
\centering
\includegraphics[width=\textwidth]{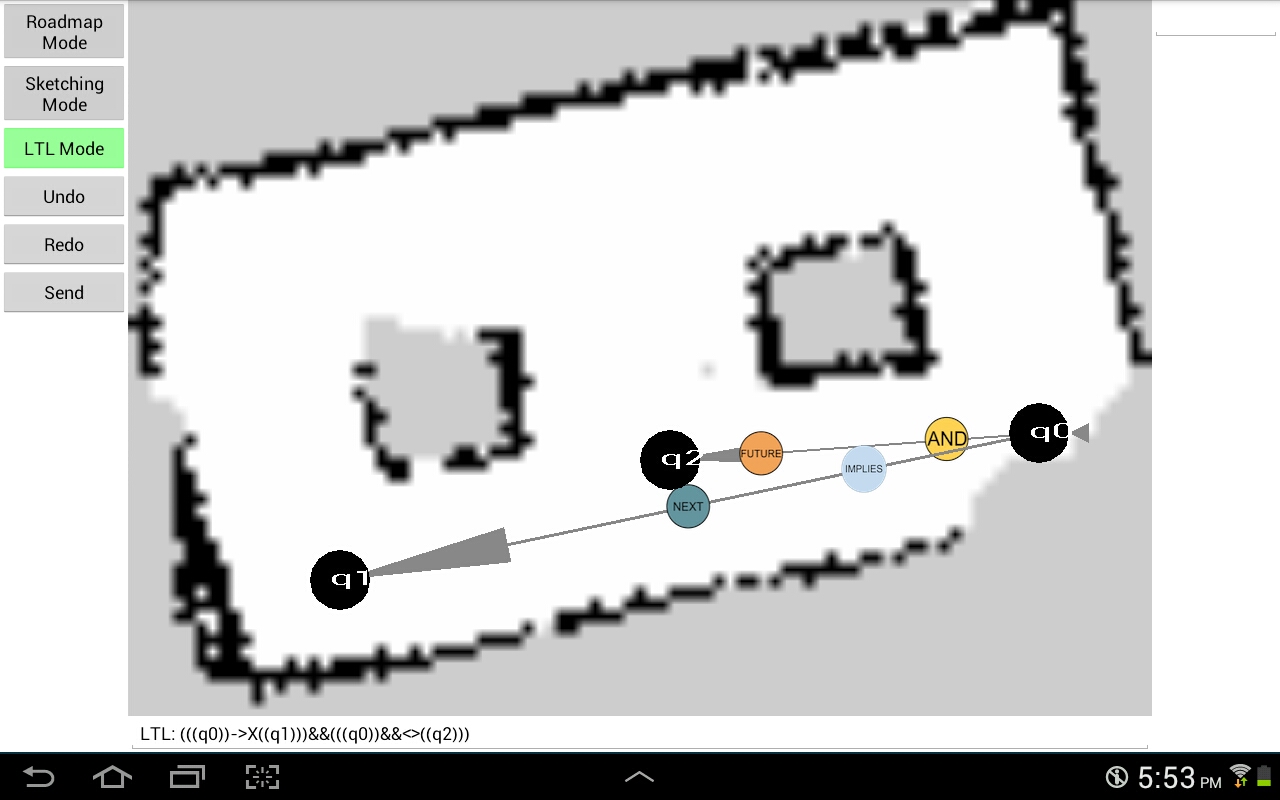}
\end{subfigure}
\hfill
\begin{subfigure}[b]{0.3\textwidth}
\centering
\includegraphics[width=\textwidth]{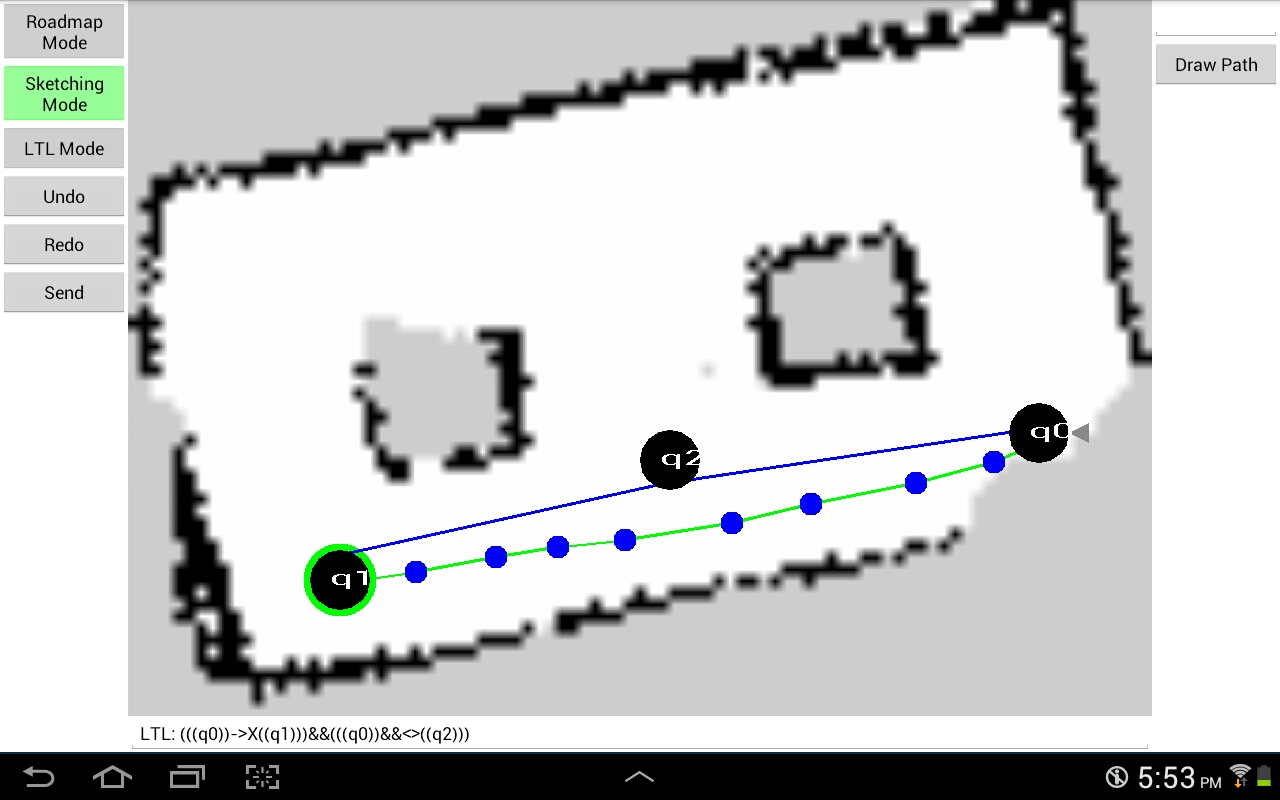}
\end{subfigure}
\hfill
\begin{subfigure}[b]{0.3\textwidth}
\centering
\includegraphics[width=\textwidth]{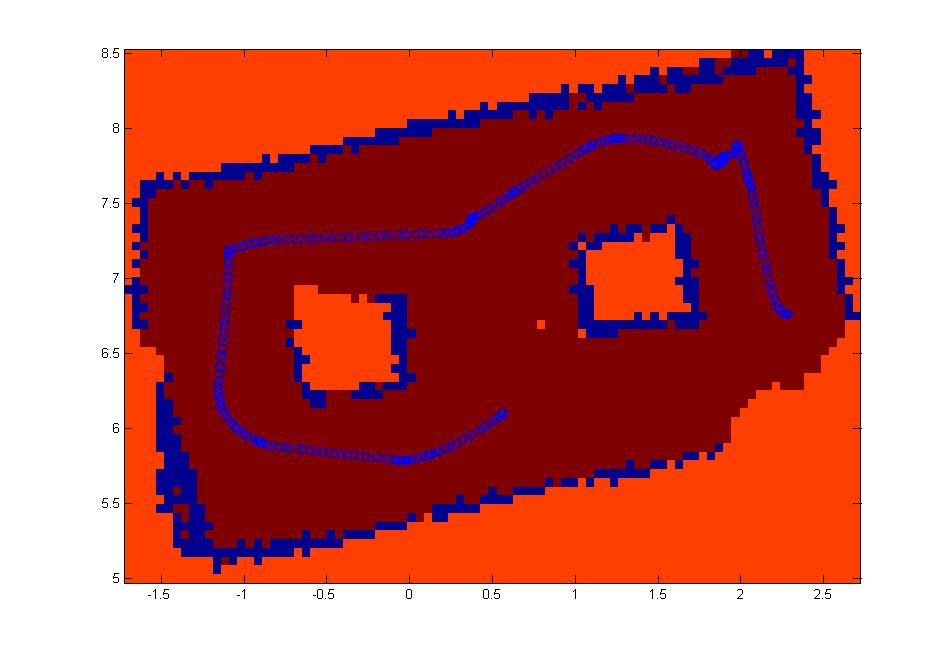}
\end{subfigure}
\caption{Experiment 1: LTL specification, sketched path and resulting trajectory.
%The specification is $(q0 \rightarrow \mathbf{X}q1) \land (q0 \land \mathbf{F}q2) \land (q0 \rightarrow \mathbf{X} \neg q2)$.}
The LTL specification is $(q0 \rightarrow \mathbf{X}q1) \land (q0 \land \mathbf{F}q2)$.}
\label{fig:ltl_spec1}
\end{figure}
}
{
\begin{figure}
\centering
\begin{subfigure}[b]{0.1583\textwidth}
\centering
\includegraphics[width=\textwidth]{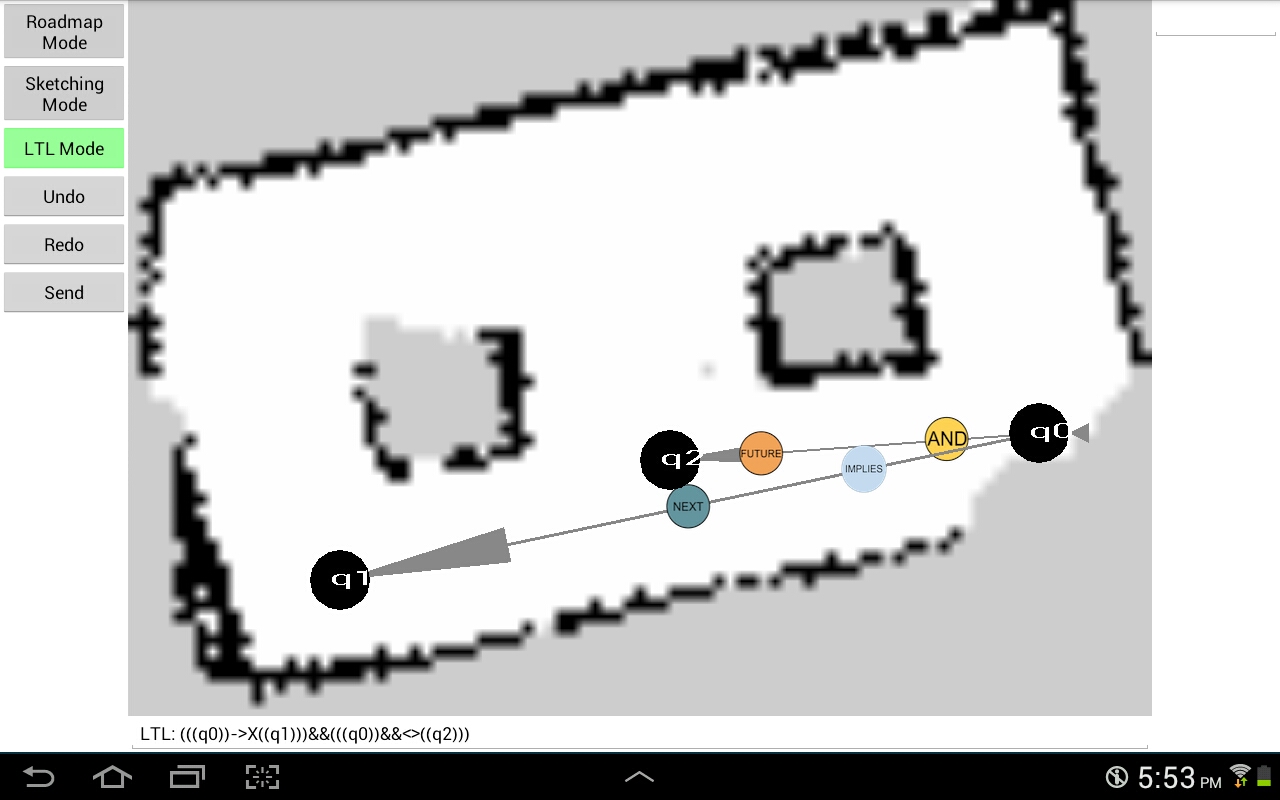}
\end{subfigure}
\hfill
\begin{subfigure}[b]{0.1583\textwidth}
\centering
\includegraphics[width=\textwidth]{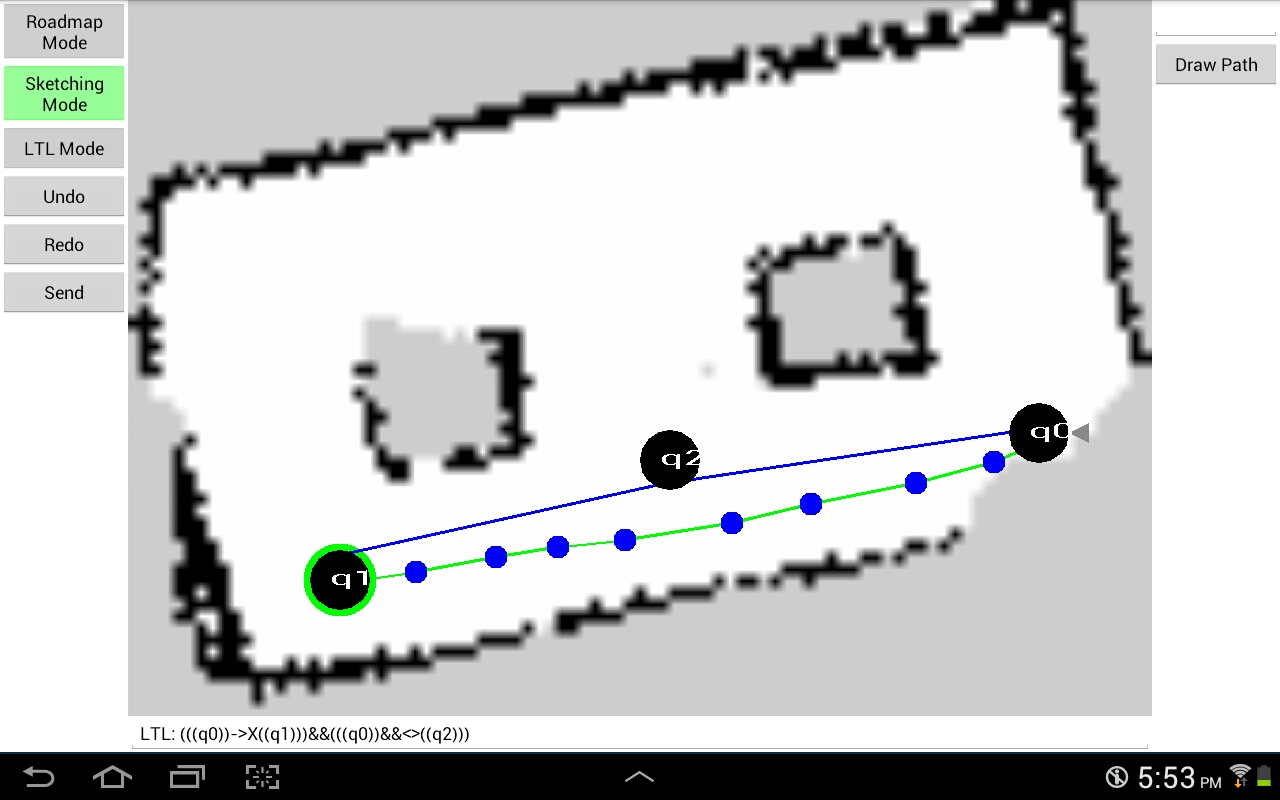}
\end{subfigure}
\hfill
\begin{subfigure}[b]{0.1583\textwidth}
\centering
\includegraphics[width=\textwidth]{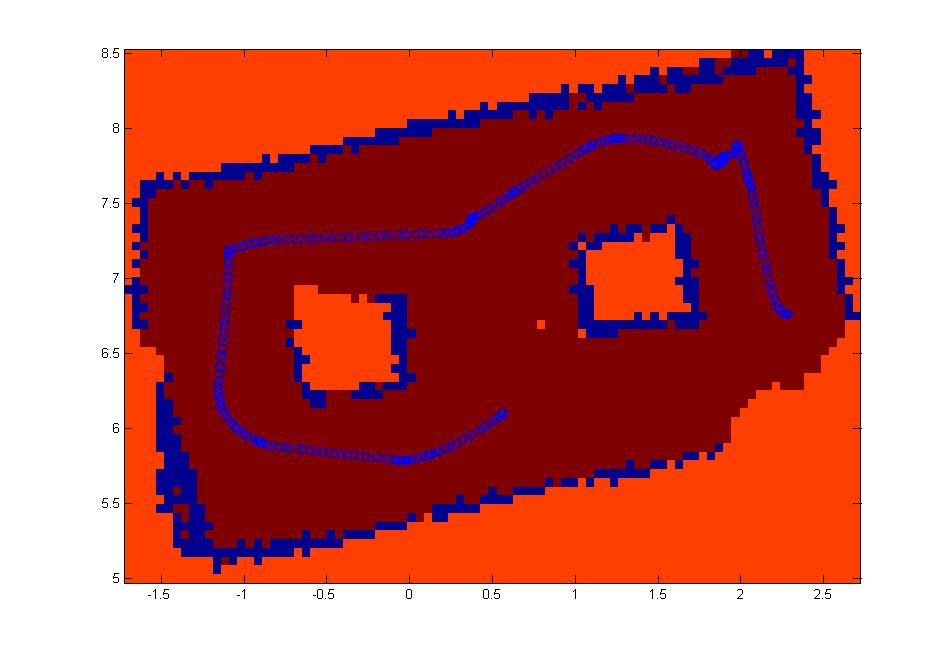}
\end{subfigure}
\caption{Experiment 1: LTL specification, sketched path and resulting trajectory.
%The specification is $(q0 \rightarrow \mathbf{X}q1) \land (q0 \land \mathbf{F}q2) \land (q0 \rightarrow \mathbf{X} \neg q2)$.}
The LTL specification is $(q0 \rightarrow \mathbf{X}q1) \land (q0 \land \mathbf{F}q2)$.}
\label{fig:ltl_spec1}
\end{figure}
}

%In the second experiment, we provided another set of conditions (shown in Fig. \ref{fig:ltl_spec2}).
Figure \ref{fig:ltl_spec2} shows the second experiment.
The task of the TurtleBot is to follow the specification $(q0 \land \mathbf{G}\mathbf{F}(q1 \land \mathbf{F}q2))$.
In natural language, it means ``the TurtleBot is required to start from q0 and head for q1 then q2 and loop between q1 and q2''.
It took 4 milliseconds in average, having M=6 and N=29.

%\todo[inline]{TODO: add Figure 9 below.}

\ifthenelse {\boolean{TECHREP}}
{
\begin{figure}
\centering
\begin{subfigure}[b]{0.3\textwidth}
\centering
\includegraphics[width=\textwidth]{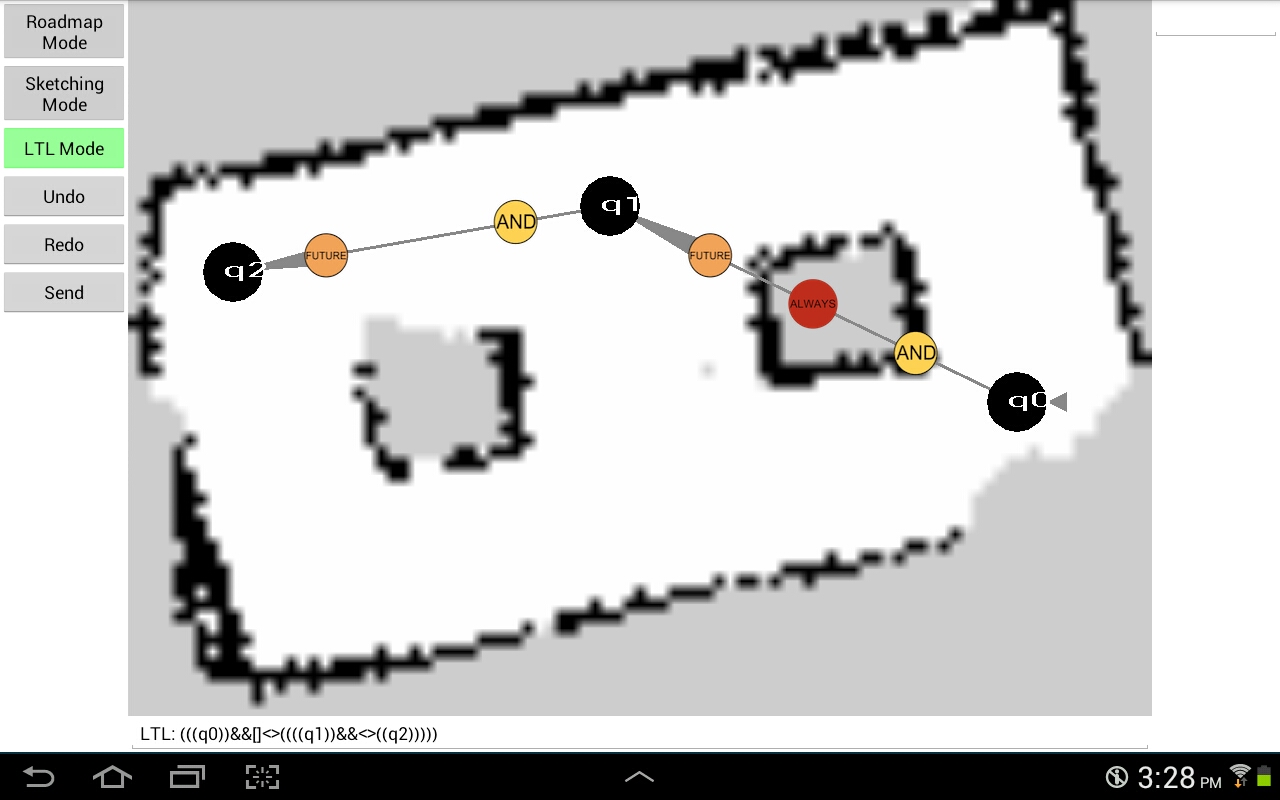}
\end{subfigure}
\hfill
\begin{subfigure}[b]{0.3\textwidth}
\centering
\includegraphics[width=\textwidth]{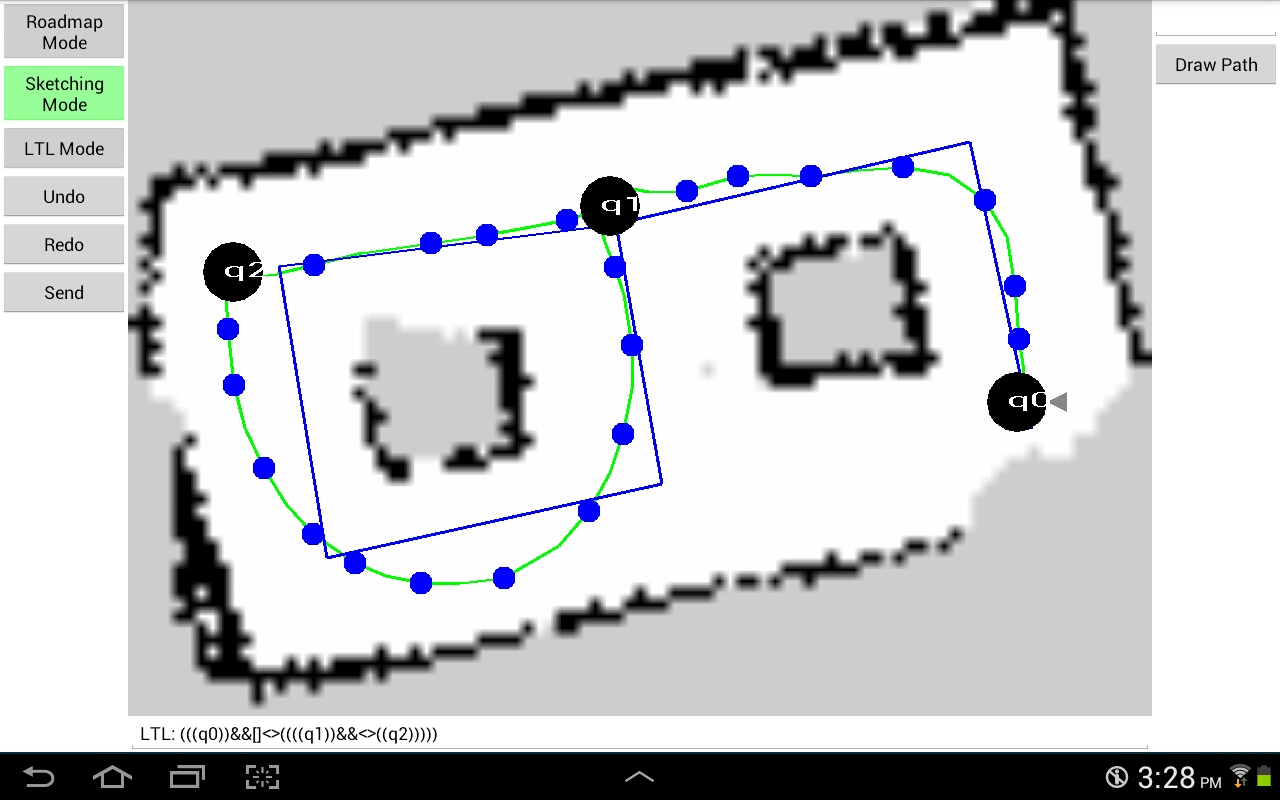}
\end{subfigure}
\hfill
\begin{subfigure}[b]{0.3\textwidth}
\centering
\includegraphics[width=\textwidth]{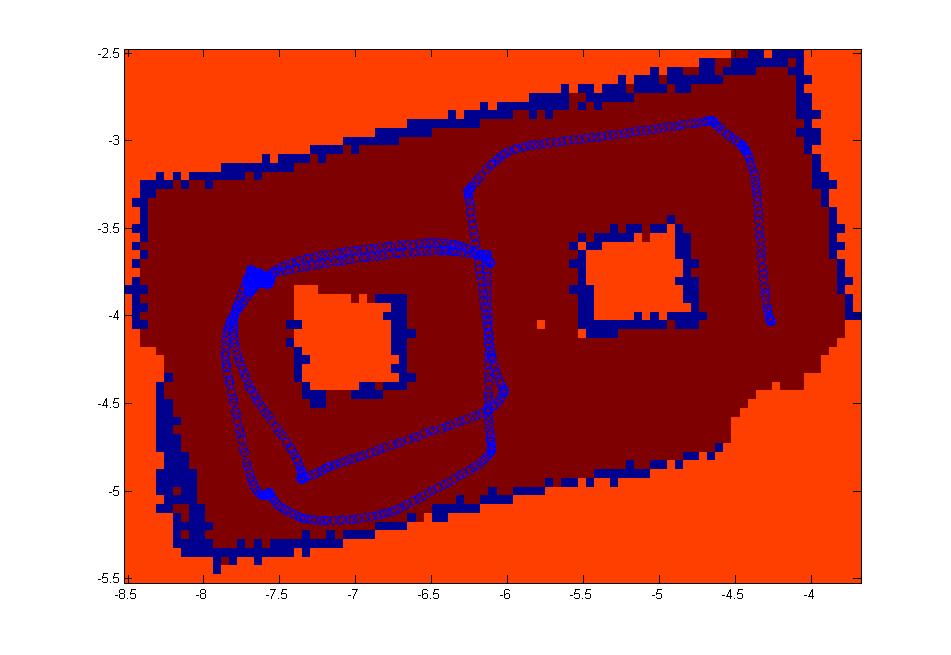}
\end{subfigure}
\caption{Experiment 2: LTL specification, sketched path and resulting trajectory. The LTL specification is  $(q0 \land \mathbf{G}\mathbf{F}(q1 \land \mathbf{F}q2))$.}
\label{fig:ltl_spec2}
\end{figure}
}
{
\begin{figure}
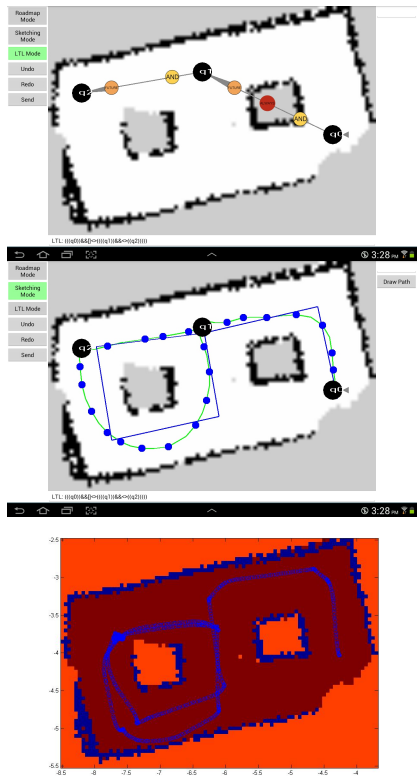

\centering
\begin{subfigure}[b]{0.1583\textwidth}
\centering
\includegraphics[width=\textwidth]{figure15_1}
\end{subfigure}
\hfill
\begin{subfigure}[b]{0.1583\textwidth}
\centering
\includegraphics[width=\textwidth]{figure15_2}
\end{subfigure}
\hfill
\begin{subfigure}[b]{0.1583\textwidth}
\centering
\includegraphics[width=\textwidth]{figure15_3}
\end{subfigure}
\caption{Experiment 2: LTL specification, sketched path and resulting trajectory. The LTL specification is  $(q0 \land \mathbf{G}\mathbf{F}(q1 \land \mathbf{F}q2))$.}
\label{fig:ltl_spec2}
\end{figure}
}

In both experiments, the TurtleBot succeeded in finding the correct path and followed the plan.
%More details on the experiment can be found in \cite{eLTLvis_wiki}.
More details on the experiment can be found in \cite{eLTLvisWiki}.

\section{Conclusions}
Our current research aims to solve the path planning problem with a path requirement under an LTL specification for a single robot.
We combined the ease of use of a sketch interface and LTLvis \cite{SrinivasKKKF2013} into a hybrid interface to allow users input customized paths. We conducted two experiments.
The interface can express user demands and the planner can realize these demands correctly in the experiments.
In terms of future research, the interface can be extended to multiple robots by adding a cooperation module.
Second, we can add a real-time feedback module to the planner so that the users will know how the robots are running.
Third, we plan to perform a usability study to test its ease of use.

\section*{Acknowledgment}
This work was partially supported by NSF CPS 1446730.

\bibliographystyle{IEEEtran}
\bibliography{IEEEabrv,eltlvis_ref}

\end{document}